\documentclass[runningheads]{llncs}
\usepackage{graphicx}

\usepackage{tikz}
\usepackage{comment}
\usepackage{amsmath,amssymb} 
\usepackage{color}

\usepackage[accsupp]{axessibility}  

\usepackage{amsmath}
\usepackage{amssymb}
\usepackage{pifont}
\newcommand{\cmark}{\ding{51}}%
\newcommand{\xmark}{\ding{55}}%

\usepackage{newfloat}
\usepackage{listings}

\usepackage{algorithm}
\usepackage{algpseudocode}
\usepackage{multirow}
\usepackage{booktabs}
\usepackage{subfigure}

\usepackage{verbatim}

\usepackage{color}
\usepackage{colortbl}

\usepackage{wrapfig,lipsum,booktabs}

\definecolor{black}{rgb}{0,0,0}
\definecolor{red}{rgb}{1,0,0}
\definecolor{blue}{rgb}{0,0,1}
\definecolor{green}{rgb}{0.2,0.8,0}

\makeatletter
\def\hlinewd#1{%
      \noalign{\ifnum0=`}\fi\hrule \@height #1 \futurelet
      \reserved@a\@xhline}

\makeatletter
\def\blfootnote{\xdef\@thefnmark{}\@footnotetext}
\makeatother

\newcommand{\figref}[1]{Fig. \ref{#1}}
\newcommand{\tabref}[1]{Table \ref{#1}}
\newcommand{\equref}[1]{(\ref{#1})}
\newcommand{\secref}[1]{Sec. \ref{#1}}
\renewcommand{\algref}[1]{Alg. \ref{#1}}
\renewcommand{\equref}[1]{Eq.~(\ref{#1})}

\newcommand{\model}{PointFix}
\newcommand{\eg}{\emph{e.g. }}
\newcommand{\ie}{\emph{i.e., }}

\begin{document}
\pagestyle{headings}
\mainmatter
\def\ECCVSubNumber{2217}  

\title{PointFix: Learning to Fix Domain Bias for Robust Online Stereo Adaptation} 


\titlerunning{PointFix: Learning to Fix Domain Bias for Robust Online Stereo Adaptation}
%
\author{Kwonyoung Kim$^{1}$ \quad
Jungin Park$^{1}$ \quad
Jiyoung Lee$^{2}$ \\
Dongbo Min$^{3}$ \quad
Kwanghoon Sohn$^{1}$\thanks{Corresponding author.}}

\institute{
$^{1}$Yonsei University \quad $^{2}$NAVER AI Lab \quad $^{3}$Ewha Womans University \\ \vspace{3pt}
{\tt\small $\lbrace$kyk12, newrun, khsohn$\rbrace$@yonsei.ac.kr} \quad \\
{\tt\small lee.j@navercorp.com} \quad
{\tt\small dbmin@ewha.ac.kr}}


%
\authorrunning{K. Kim et al.}
%

\maketitle

\begin{abstract}
    \blfootnote{{\small
    \noindent This work was supported by the National Research Foundation of Korea(NRF) grant funded by the Korea government(MSIP) (NRF-2021R1A2C2006703) and the Yonsei University Research Fund of 2021 (2021-22-0001).}}
    Online stereo adaptation tackles the domain shift problem, caused by different environments between synthetic (training) and real (test) datasets, to promptly adapt stereo models in dynamic real-world applications such as autonomous driving.
    However, previous methods often fail to counteract particular regions related to dynamic objects with more severe environmental changes.
    To mitigate this issue, we propose to incorporate an auxiliary point-selective network into a meta-learning framework, called \textit{PointFix}, to provide a robust initialization of stereo models for online stereo adaptation.
    In a nutshell, our auxiliary network learns to fix local variants intensively by effectively back-propagating local information through the meta-gradient for the robust initialization of the baseline model.
    This network is model-agnostic, so can be used in any kind of architectures in a plug-and-play manner. We conduct extensive experiments to verify the effectiveness of our method under three adaptation settings such as short-, mid-, and long-term sequences. Experimental results show that the proper initialization of the base stereo model by the auxiliary network enables our learning paradigm to achieve state-of-the-art performance at inference.
\end{abstract}

\section{Introduction}
    \vspace{-10pt}
    Stereo depth estimation to predict 3D geometry for practical real-world applications such as autonomous driving~\cite{achtelik2009stereo} has been developed by handcrafted methods~\cite{sgm,geiger2010efficient,zhang2009cross,hu2012quantitative} and deep stereo models based on supervised learning~\cite{dispnet,psmnet,tankovich2021hitnet,khamis2018stereonet} that leverage the excellent representation power of deep neural networks.
    In general, given that the high performance of deep networks is guaranteed when test and training data are derived from a similar underlying distribution~\cite{finn2017model,liu2016coupled,hoffman2016fcns,online_meta_MSSS,pmlr-v97-finn19a}, they demand a huge amount of annotated training data to reflect a real-world distribution.
    Acquiring groundtruth disparity maps, but unfortunately, is laborious and impractical~\cite{tonioni2019learning}.
    Especially for autonomous driving, constructing datasets from all possible different conditions (\eg weather and road conditions) is impossible while it is a very fatal problem~\cite{tonioni2019real}.
    To mitigate the aforementioned issues, an intuitive solution is to finetune the stereo model trained on a large-scale synthetic dataset that is easier to collect groundtruth.
    However, despite the help of large-scale synthetic datasets, most recent works~\cite{tonioni2017unsupervised,pang2018zoom,zhou2017unsupervised,zhang2018activestereonet} have pointed out the limitation of fine-tuning that is incapable of collecting sufficient data in advance when running the stereo models in the \textit{open world}.
    While domain generalization methods~\cite{dsmnet,cfnet} have shown promising results without real images, they require high computations to provide generalized stereo models and often fail to respond to continuously changing environments.

    \begin{figure}[t]
       \centering
        \renewcommand{\thesubfigure}{}
       \subfigure[(a) Input image]{\includegraphics[width=0.8 \linewidth]{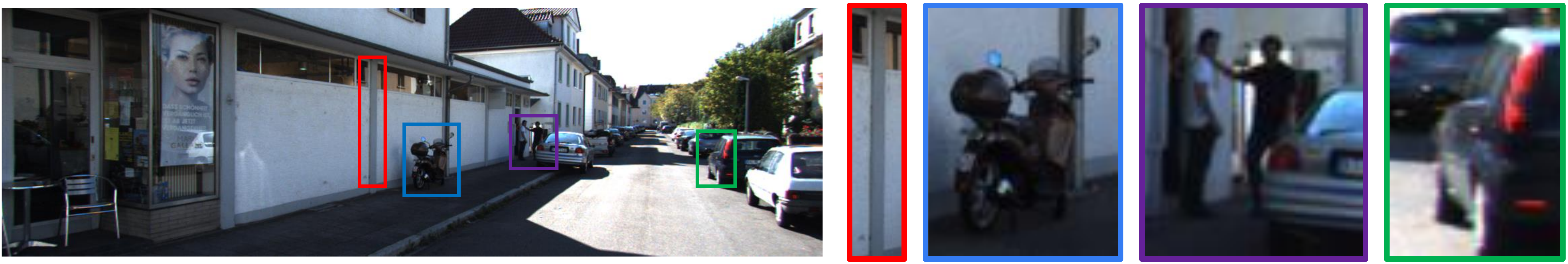}} \\\vspace{-7pt}
       \subfigure[(b) Disparity and reprojection error of MADNet]{\includegraphics[width=0.8 \linewidth]{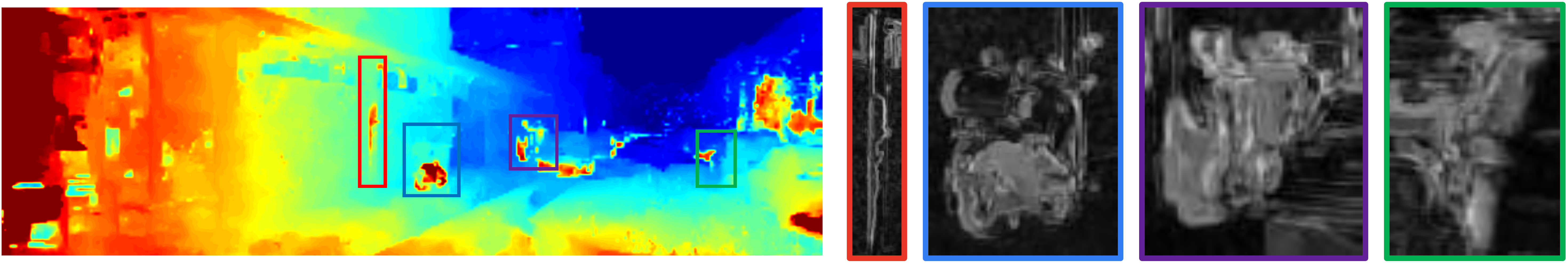}} \\\vspace{-7pt}
       \subfigure[(c) Disparity and reprojection error of \model~(ours)]{\includegraphics[width=0.8 \linewidth]{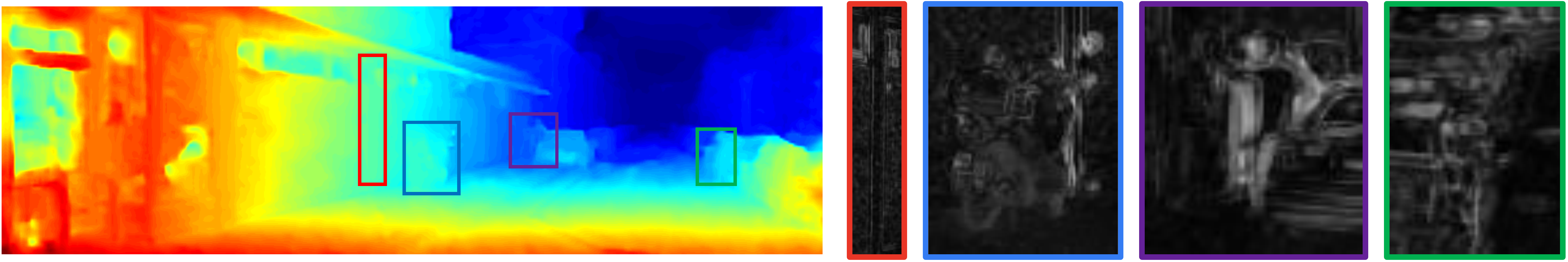}}\vspace{-7pt}
       \caption{Estimated disparity after online adaptation from MADNet and our \model. Our method has a much stronger adaptation ability, especially in local detail.}\vspace{-10pt} \label{fig:1}
    \end{figure}

    As an alternative solution, online stereo adaptation~\cite{tonioni2019real,tonioni2019learning,wang2020faster,poggi2021continual,zhang2019online,knowlestoward} is proposed to incorporate unsupervised domain adaptation~\cite{garg2016unsupervised,godard2017unsupervised} into a continual learning process~\cite{rusu2016progressive}.
    Formally, a baseline network is trained offline using a large number of the labeled synthetic datasets (\eg Synthia~\cite{ros2016synthia}, FlyingThings3D~\cite{dispnet}) and continually adapted to unlabeled unseen scenarios at test time in an unsupervised manner.
    To demonstrate a faster inference speed for real-world applications,
    MADNet~\cite{tonioni2019real,poggi2021continual} proposed a lightweight network and modular adaptation framework to rely on self-supervision via reprojection loss~\cite{godard2017unsupervised}.
    Meanwhile, Learning-to-adapt (L2A)~\cite{tonioni2019learning} introduced a new learning framework based on model agnostic meta-learning (MAML)~\cite{finn2017model} for the improved adaptation ability of the network by well-suited base parameters.
    It shows that the meta-learning framework has great potential in making the network parameters in learning process to make the parameters into a very adaptable state.
    Despite their great progress, most online adaptation methods~\cite{tonioni2019learning,tonioni2019real} have merely attempted to only impose a global average errors from the whole prediction as a learning objective during an offline training without attention to a domain gap in local, showing poor initial performance.

    In particular, we observe that given stereo images from a novel environment, incorrectly estimated disparities are concentrated on specific \textit{local} regions, as depicted in~\figref{fig:1}(b).
    The domain shift~\cite{patricia2017deep} issue arises because the local context of test data (\eg appearance deformations of objects, occlusion type, or the form of a shadow etc.) is significantly diverse from those deployed throughout the training process.
    This means that without taking such locally varying discrepancies between training and test data into account, the global adaptation strategy used in the existing methods~\cite{tonioni2019real,tonioni2019learning,wang2020faster,poggi2021continual,zhang2019online,knowlestoward} has fundamental limitations in improving the adaptation performance.
    In addition, a plug-and-play algorithm to easily combine with evolving deep stereo networks is also needed.

    In this paper, we propose a novel model-agnostic training method for robust online stereo adaptation, called \textit{\model}, that can be flexibly built on the top of existing stereo models and learns the base stereo network on a meta-learning framework.
    Unlike the existing methods~\cite{poggi2021continual,knowlestoward,chen2020consistency,wang2020faster} that focus on a new online adaptation strategy, we leverage the meta-learning strategy for \textit{learning-to-fix} a base stereo network offline so that it can have generalized initial model parameters and respond to novel environments more robustly.
    Specifically, we incorporate an auxiliary point-selective network, termed \model Net into meta-learning to rectify the local detriment of the base network and alternately fix the base network by an additional update as in the online meta-learning methods~\cite{online_meta_MSSS,pmlr-v97-finn19a}.
    As a result, the parameters of the base network are updated to grasp and utilize the incoming local context and can be robust to the local variants at test time by preventing the network from being biased to global domain dependencies only.
    
    In the experiment, we learn two base stereo models, DispNetC~\cite{dispnet}, MADNet~\cite{tonioni2019real}, together with our framework on the synthetic data in the offline training, and then update the whole models (full adaptation) or the sub-module of the models (MAD adaptation) in the online adaptation using the unsupervised reprojection loss~\cite{garg2016unsupervised,godard2017unsupervised} on the real-world dataset.
    Note that the proposed auxiliary network is not used in the online adaptation during inference, maintaining an original inference speed of the base stereo models.
    Given that our \model~is a general and synergistic strategy that can be adopted with any kinds of stereo networks, it improves a generalization capability of the base stereo model to novel environments through the robust parameter initialization. 
    Extensive experimental results show that \model~outperforms recent state-of-the-art results by a significant margin on various adaptation scenarios including short-, mid-, and long-term adaptation. In addition, comparison with domain generalization methods~\cite{dsmnet,cfnet,lipson2021raft} demonstrates the superiority of our \model~in terms of both accuracy and speed.\vspace{-5pt}

\section{Related Work}
\vspace{-7pt}
\begin{figure*}[!t]
    \begin{center}
       \includegraphics[width=0.96\linewidth]{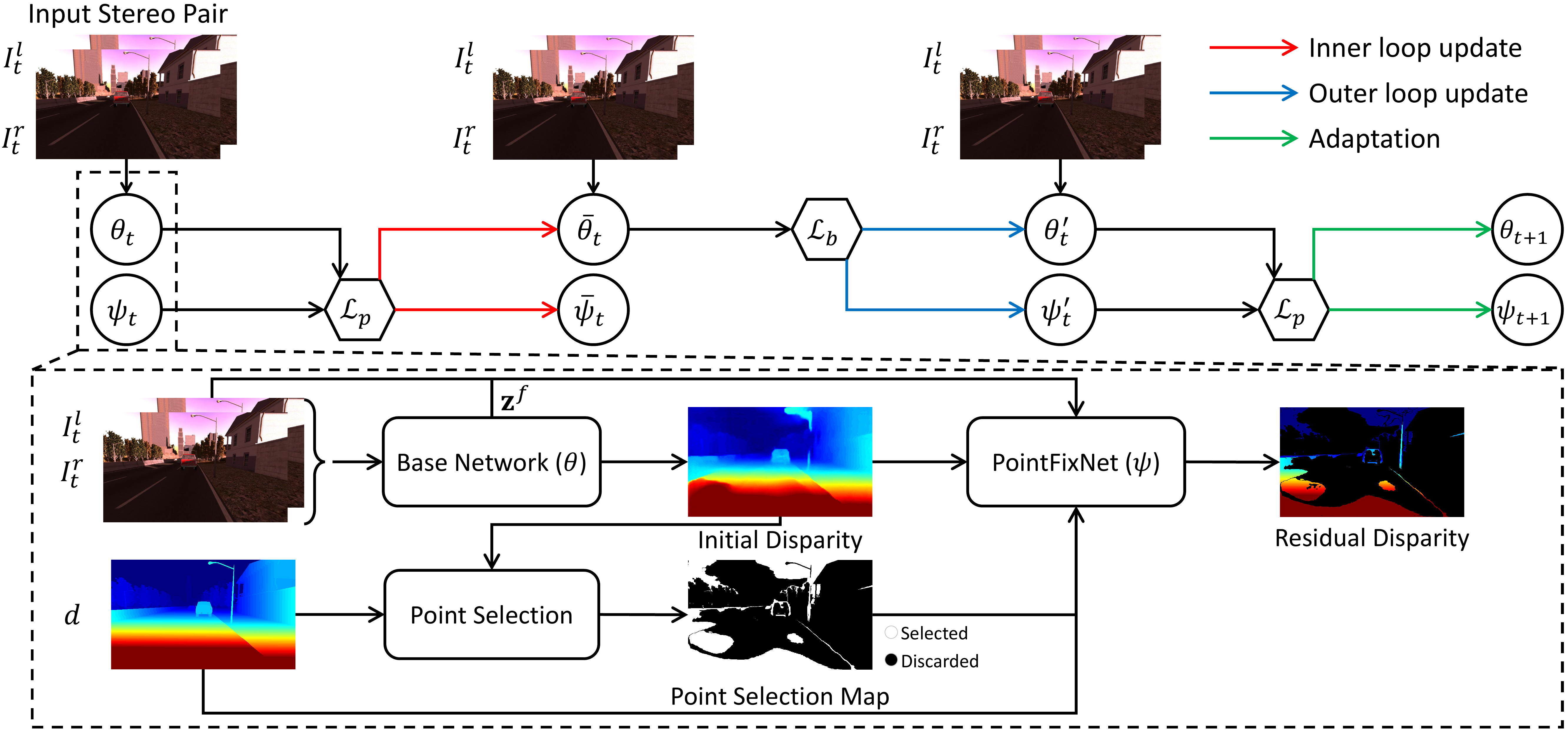}
    \end{center}\vspace{-20pt}
       \caption{
       The overall framework of the proposed \textit{PointFix} and \model Net. We alternatively update the base network and \model Net underlying meta-learning. The detail of main flowchart is illustrated in the dotted box. 
       }\vspace{-10pt}
    \label{fig:2}
\end{figure*}

For depth estimation from stereo images, there is an extensive literature, but here we briefly introduce related work in the application of convolutional neural networks (CNNs).
Modern approaches using CNNs are mostly categorized by matching-based approaches~\cite{zbontar2016stereo,khamis2018stereonet,sun2018pwc} that learn how to match corresponding points, and regression-based methods~\cite{gcnet,psmnet,dispnet} that learn to directly regress sub-pixel disparities. 
To further enhance the performance, some works~\cite{kim2018unified,kim2019laf,poggi2021conf} consider exploiting auxiliary network or module to assist base network by estimating confidence of prediction map from stereo inputs, prediction or cost volume of the base network.
Although their results are promising, they have a limitation in retaining the superb accuracy in new domains~\cite{poggi2021continual}.

Pointing out the domain shift issue, recent works~\cite{tonioni2019learning,tonioni2019real,poggi2021continual,zhang2019online,wang2020faster,chen2020consistency,knowlestoward} have proposed online stereo adaptation methods to consider more practical solution for real-world applications.
They have argued that we should consider a new \textit{open-world} scenario in which input frames are sequentially provided to the model with certain time intervals.
By continually updating parameters at test time in an unsupervised manner, they observe the adaptability of the models in changing environments.
As one of the pioneering works, \cite{tonioni2019real,poggi2021continual} has proposed a light weight model (MADNet) and its modular updating frameworks (MAD, MAD++) to improve the adaptation speed drastically.
Following this idea, \cite{wang2020faster} has picked up the speed even more by implementing `Adapt or Hold' mechanism based on deep Q-learning network~\cite{mnih2015human}.

Closely related work to ours is meta-learning based online adaptation approaches~\cite{tonioni2019learning,zhang2019online} that train to learn proper model parameters to be better suitable for online adaptation.
L2A~\cite{tonioni2019learning} has directly incorporated online adaptation process into the inner loop updating process, and also learn a confidence measure to use adaptive weighted loss.
However, L2A is inherently unstable during training due to multiple adaptation steps, especially when coupled with the lightweight model (\eg MADNet~\cite{poggi2021continual}).
Also, it is worth noting a difference in purpose between confidence-weighted adaptation of L2A and our \model Net.
The role of the auxiliary network in L2A is to eliminate uncertain errors from inherently noisy reprojection loss, whereas ours is to stabilize the generality of the base model by generating a proper point-wise learning objective for particular bad pixels.
We hereby deliver the gradient of the point-wise loss to the base stereo model to prevent the model from being learned by minimizing only the global errors and to remedy domain-invariant representations in local regions. \vspace{-5pt}

\section{Problem Statement and Preliminaries}
 \vspace{-3pt}  
    \subsubsection{Online Stereo Adaptation}
    Here we first formulate online stereo adaptation.
    Given stereo image pairs from source domain $\mathcal{D}_{s}$ with available ground-truth maps, online stereo adaptation aims to learn a stereo model capable of adapting itself dynamically in a novel unseen domain $\mathcal{D}_{u}$.
    In the inference, given a set of parameters $\theta$ from the base stereo model trained on $\mathcal{D}_{s}$, the base network parameters are updated in a single iteration step $t$ to adapt the stereo model with respect to continuous input sequence without ground-truth disparity maps:
    \begin{equation}\label{equ:1}
        \theta_{t+1} \leftarrow \theta_{t} - \alpha\nabla_{\theta_t}\mathcal{L}_{u}(\theta_t, I^{l}_{t}, I^{r}_{t}),
    \end{equation}
    where $\mathcal{L}_{u}$ is an unsupervised loss function, $\alpha$ is the learning rate, $I^{l}_{t}$, and $I^{r}_{t}$ are left and right images of the $t$-th stereo pair from $\mathcal{D}_{u}$.
    We evaluate the performance of the online adaptation under the short-term (sequence-level), mid-term (environment-level), and long-term (full) settings.
    Note that following previous approaches~\cite{tonioni2019learning,tonioni2019real}, we use a reprojection error~\cite{godard2017unsupervised} as the unsupervised learning objective $\mathcal{L}_{u}$.

\vspace{-10pt}
\subsubsection{Model Agnostic Meta-learning}
    MAML~\cite{finn2017model} has proposed to learn initial base parameters that are suited to adapt to new domain with only few updates.
    This is attained by implementing a nested optimization which consists of an inner loop and an outer loop.
    Specifically, the inner loop updates the base parameters for each sample in batch separately in the standard gradient descent way.
    The outer loop performs the update of the base parameters using the sum of sample-specific gradients (meta-gradient) which are computed by the parameters updated in the inner loop.
    Formally, for a set of tasks $\mathcal{T}$, let the meta-training and meta-testing sets be $\mathcal{D}_{\tau}^{\text{train}}$ and $\mathcal{D}_{\tau}^{\text{test}}$ respectively.
    A set of parameters $\theta^*$ can be obtained for a specific task $\tau\in\mathcal{T}$ with a single gradient step in the inner loop:
    \begin{equation}
        \theta^* = \min_{\theta}\sum_{\tau\in\mathcal{T}}{\mathcal{L}(\theta - \alpha\nabla_{\theta}\mathcal{L}(\theta, \mathcal{D}_\tau^{\text{train}}), \mathcal{D}_{\tau}^{\text{test}})},
    \end{equation}
    where $\mathcal{L}$ is a objective function for the task and $\alpha$ is the learning rate.
    They carries out meta-update, including inner and outer updates, during training and then adaptation for the target task after the the whole meta-learning processes are completely over.
    In this paper, we treat the disparity prediction for each stereo pair as a single task.
    
    Recent works \cite{online_meta_MSSS,pmlr-v97-finn19a} have introduced two stage meta-learning  procedure (a.k.a online meta-learning) that implements meta-update and adaptation stages alternatively during training to achieve more efficient gradient path and better performance.
    Inspired by their alternative updating scheme, we propose a novel \textit{learning-to-fix} strategy.
    Our objective is to get base network and \model Net parameters that enable the base network to quickly adapt regardless of local variants in unseen domains. 
    The details are described in the following section.

\section{PointFix: Learning to Fix}\vspace{-5pt}
    We design a novel meta-learning framework for online stereo adaptation to learn good base model and quickly adapt to novel environments (\ie unseen domain), especially concentrating on erroneous pixels.
    As illustrated in \figref{fig:2}, we leverage off-the-shelf deep stereo model as a base network and incorporate an auxiliary network, \model Net, to make parameters of the base network robust to the local distortion.
    
    \begin{algorithm}[t]
        \caption{Parameter update with \model~loss}\label{alg:1}
        \textbf{Input}: $\mathcal{I}=\{(I^{l}_n, I^r_n, d_n)\}_{n=0}^{N-1}$; learning rate $\alpha$; base model parameters $\theta$; PointFix network parameters $\psi$\\
        \textbf{Output}: updated parameters ~$\bar{\theta}, ~\bar{\psi}$
        \begin{algorithmic}[1]
        \Function{PointUpdate}{$\mathcal{I}, \alpha, \theta, \psi$}
        \State {$\mathcal{L}_{p}^{\tau} \leftarrow 0$}  \Comment{Initialize loss}
        \For {$n=0$ \textbf{to} $N-1$} 
        \State {$\hat{d}, \mathbf{z}^{b} = \mathcal{F}(I^{l}_n, I^{r}_n | \theta)$}   
        \State {$\mathbf{p}(\theta) = \{(i,j)|~|(\hat{d}_{ij}-d_{n, ij})|_1 > 3\}$}   
        \Comment{Select points}
        \State {$\mathbf{z}^{c} = \mathcal{F}(I^{l}_n, \hat{d}, d_n | \psi^c)$}   
            \For {$(i,j)\in\mathbf{p}(\theta)$}
                \State {$r_{ij} = \mathcal{F}(\mathbf{z}^{b}_{ij}, \mathbf{z}^{c}_{ij} | \psi^p)$}    \Comment{Residual disparity}
                \State {$\mathcal{L}_{p}^{\tau} \leftarrow \mathcal{L}_{p}^{\tau} + \mathcal{L}_{p}(r_{ij}+\hat{d}_{ij}, d_{n, ij})$} 
                \Comment{\model~loss}
            \EndFor
        \EndFor
        \State{$\bar{\theta} \leftarrow \theta - \alpha {\nabla_{\theta} \mathcal{L}_{p}^{\tau}}$} \Comment{Update base network}
        \State{$\bar{\psi} \leftarrow \psi - \alpha {\nabla_{\psi} \mathcal{L}_{p}^{\tau}}$} \Comment{Update \model Net}
        \State \textbf{return} $\bar{\theta}, \bar{\psi}$
        \EndFunction
        \end{algorithmic}  
    \end{algorithm}

    \begin{algorithm}[t]
        \caption{Overall training procedure}\label{alg:2}
        \textbf{Hyperparameters}: batch size $N$; max iteration $K$; learning rate of inner and outer loop $\alpha$, $\beta$\\
        \textbf{Input}: pre-trained base model parameters $\theta$; source training dataset $\mathcal{S}$\\
        \textbf{Output}: optimized base model parameters $\theta^*$
        \begin{algorithmic}[1]
        \Function{Training}{$\theta, \mathcal{S}$}
        \State {Initialize $\theta$ and $\psi$.}
        \For { $k = 0$ \textbf{to} $K-1$}
            \State $\mathcal{L}_k \leftarrow 0$ \Comment{Initialize loss}
            \State {$\mathcal{I}_k \sim \mathcal{S}$} \Comment{Sample a batch of size $N$}
            \For { $n = 0$ \textbf{to} $N-1$}   
                \State $\theta_n \leftarrow \theta, \psi_n \leftarrow \psi$   \Comment{Copy parameters}
                \State $\bar{\theta}_n, \bar{\psi}_n, \leftarrow \text{PointUpdate}(\mathcal{I}_{k,n}, \alpha, \theta_n, \psi_n)$ \Comment{Inner loop update}
                \State {$\hat{d}_{n}, \textbf{z}^{b}_{k,n} = \mathcal{F}(I^{l}_{k,n}, I^{r}_{k,n}|\bar{\theta}_n)$}
                \State $\mathcal{L}_k \leftarrow \mathcal{L}_k + \mathcal{L}_b(\hat{d}_{n}, d_{k,n})$  \Comment{Base loss}
            \EndFor
            \State{$\theta' \leftarrow \theta - \beta {\nabla_{\theta} \mathcal{L}_k}$}  \Comment{Outer loop update}
            \State{$\psi' \leftarrow \psi - \beta {\nabla_{\psi} \mathcal{L}_k}$}    \Comment{Outer loop update}
            \State $\theta, \psi \leftarrow \text{PointUpdate}(\mathcal{I}_k ,\alpha, \theta', \psi')$ \Comment{Adaptation}
        \EndFor
        \State \textbf{return} $\theta^* \leftarrow \theta$
        \EndFunction
        \end{algorithmic} 
    \end{algorithm}
\vspace{-7pt}
    \subsection{Base Stereo Models}
    Our goal is to train base models offline to be more suitable for the online adaptation by correcting bias to the seen domain.
    We have employed two stereo networks as a base model: 1) DispNet-Corr1D~\cite{dispnet} (shortened as DispNetC) and 2) MADNet~\cite{tonioni2019real}.
    Besides taking the initial disparity $\hat{d}$ estimated from the base model, we extract intermediate features that is useful to exploit the fine-grained information. 
    The intermediate features consist of the matching cost $\mathbf{c}$ (same as correlation layer in DispnetC) and its corresponding left feature $\mathbf{f}^l$. They are concatenated and taken as a base feature $\mathbf{z}^{b}= \Pi(\mathbf{c}, \mathbf{f}^l)$, where $\Pi(\cdot,\cdot)$ is a concatenation operation.
    We note that in MADNet, the matching cost calculation is similar with the one in DispNetC, but before calculation their right features are warped with a disparity map on a coarse resolution to reduce search range and computation. 
    However, to extract base features, we apply the same matching cost computation scheme used in DispNetC regardless of the base network to ensure the generality of our method.

    In each inner loop of our method, we select a pixel $i,j$ to fix local deformations by computing $\ell_1$ loss between $\hat{d}$ and the ground-truth $d$, such that the set of points $\mathbf{p}(\theta)$ with the base parameter $\theta$ can be derived as:
    \begin{equation}\label{equ:p}
        \mathbf{p}(\theta) = \{(i,j) |~|\hat{d}_{ij} - d_{ij}|_1 > 3\}.
    \end{equation}
    Note that we represent $\mathbf{p}$ as a function of $\theta$ to indicate that the selected point varies depending on $\theta$ updated in the learning procedure.
    In next section, we describe a way to leverage the set of points for correcting local distortions caused by the seen domain bias, depicted in~\figref{fig:1}.

\vspace{-7pt}
    \subsection{\model Net}\label{sec:42}
    To mitigate the seen domain bias of the base network, we deploy an additional auxiliary network, called \model Net, which individually repairs a disparity by incurring a proper point-wise gradient.
    The \model Net consists of two modules:
    a feature extraction module (parameterized by $\psi^c$) that extracts feature $\mathbf{z}^c$ from heterogeneous inputs;
    and a point-wise prediction module (parameterized by $\psi^{p}$) that generates residual disparity value of each point and back-propagates the point-wise errors.
    Specifically, the feature extraction module consists of three convolution layers and takes the left image, $I^l$, the initial disparity $\hat{d}$, and ground-truth $d$ as inputs to integrate context around each erroneous pixel.
    Therefore, the feature $\mathbf{z}^c$ can be obtained as follows:
    \begin{equation}
        \mathbf{z}^c = \mathcal{F}(I^l, \hat{d}, d | \psi^{c}),
    \end{equation}
    where $\mathcal{F}$ is a feed-forward process.
    
    Then, the base feature $\mathbf{z}^{b}$ from the base network and feature $\mathbf{z}^{c}$ from the feature extraction module are concatenated and fed into the point-wise prediction module to generate the residual disparity value $r_{ij}$.
    Inspired by structure in ~\cite{kirillov2020pointrend}, the module consists of four fully-connected (FC) layers to produce a single value for each pixel, such that:
    \begin{equation}
        r_{ij} = \mathcal{F}(\mathbf{z}^{b}_{ij}, \mathbf{z}^{c}_{ij} | \psi^{p}),
    \end{equation}
    where $(i.j) \in \mathbf{p}(\theta)$.
    The final disparity for $(i,j)$-th pixel is obtained by adding $r_{ij}$ to $\hat{d}_{ij}$.
    We note that FC layers share weights across all selected points.

\vspace{-7pt}
\subsection{Learning to Fix} \label{l2f}
    The key idea underlying our framework is iterating \textit{learning how to fix} first and then \textit{fixing} alternatively. First, the parameters of the \model Net learn how to generate a proper gradient to the base model in a point-wise manner such that the base model can be improved with less domain bias and then secondly the base model is updated by the learned \model Net.
    This strategy is essential because if we keep network training only with point loss, the performance of prediction after \model Net can be guaranteed but the one after the base model may not. 
    Thus, to enhance maximal performance of the base network, it is necessary to employ the alternative meta-learning structure.
    To this end, we deploy two loss functions: a base loss, $\mathcal{L}_b$, derived from the whole disparity map predicted by the base network and a point loss, $\mathcal{L}_p$, applied to the final disparity values.
    They are alternatively optimized to update $\theta$ and $\psi$ by relying on the two-stage meta-learning scheme~\cite{online_meta_MSSS,pmlr-v97-finn19a}, as described in \figref{fig:2} and \algref{alg:2}.
    
    In the inner loop, parameters are copied for each sample in batch, $\theta_n\leftarrow\theta$ and $\psi_n\leftarrow\psi$. Then we calculate the \model~loss, to evaluate of the current parameters:
    \begin{align}
        \bar{\theta}_n \leftarrow \theta_n - \alpha \nabla_{\theta_{n}} \sum\nolimits_{(i,j)\in\mathbf{p}(\theta)} &\mathcal{L}_{p}(\hat{d}_{ij} + r_{ij}, d_{ij}),    \\
        \bar{\psi_n} \leftarrow \psi_n - \alpha \nabla_{\psi_n} \sum\nolimits_{(i,j)\in\mathbf{p}(\theta)} &\mathcal{L}_{p}(\hat{d}_{ij} + r_{ij}, d_{ij}),
    \end{align}
    where $\alpha$ is a learning rate, $\mathbf{p}({\theta}$) is a set of selected points, and $\mathcal{L}_{p}$ is a point-wise $\ell_1$ loss between the final disparity and its corresponding ground-truth.
    Given in \algref{alg:1}, since the \model~loss is imposed on the local distortion of the erroneous pixels selected on the initial prediction, we can update base parameters that refer to the fine-grained details.

    In the outer loop, we evaluate the performance of the updated base parameter after the inner loop.
    To measure the performance, we apply the conventional supervised loss between the initial disparity map and ground-truth.
    Following the procedure of \cite{finn2017model}, the parameters $\theta$ and $\psi$ are updated based on sum of $\mathcal{L}_{k}$ as follows:
    \begin{equation}
        \theta' \leftarrow \theta - \beta \nabla_{\theta} \mathcal{L}_k,~~~
        \psi' \leftarrow \psi - \beta \nabla_{\psi} \mathcal{L}_k,
    \end{equation}
    where $\mathcal{L}_{k} = \sum_{N} \mathcal{L}_{b}(\hat{d}_{n}, d_{k,n})$ and $k$ is the current iteration step.
    Note that the gradients are computed along with the parameters before being updated in the inner loop.
    
    Unlike traditional MAML where the parameters are optimized via meta-update only, we deploy an additional update inspired by online meta-learning.
    $\theta'$ and $\psi'$ are updated in the same way as the inner loop so that the parameters after final update, $\theta$ and $\psi$ can be written as:
    \begin{align}
        \theta \leftarrow \theta' - \alpha {\nabla_{\theta'} \sum\nolimits_{(i,j)\in\mathbf{p}(\theta')} \mathcal{L}_{p}(\hat{d}_{ij} + r_{ij}, d_{ij})},    \\
        \psi \leftarrow \psi' - \alpha {\nabla_{\psi'} \sum\nolimits_{(i,j)\in\mathbf{p}(\theta')} \mathcal{L}_{p}(\hat{d}_{ij} + r_{ij}, d_{ij})},
    \end{align}
    where $r_{ij} = \mathcal{F}(\mathbf{z}_{ij}^{b}, \mathbf{z}_{ij}^{c}, d_{ij} | \bar{\psi}^{p})$ for $(i,j)\in \mathbf{p}(\theta')$.
    Finally, the networks are updated with \model~loss which are, at the first stage, trained to generate proper back-propagation to enhance the performance of the base network in the next training step.
    At test time, we use the final parameters of the base network $\theta^*$ and perform adaptation according to \equref{equ:1}. \vspace{-5pt}

\section{Experiments}
\vspace{-5pt}

\subsection{Experimental Settings}
    \vspace{-3pt}
    \subsubsection{Datasets.}
    In order to evaluate our method on realistic scenario, we use synthetic dataset for offline training and real dataset for the test.
    Therefore, the training and test data exist in completely different data distributions. 
    Following the previous work~\cite{tonioni2019learning}, we train our networks using the Synthia~\cite{ros2016synthia} dataset and evaluate each model on the KITTI-raw~\cite{geiger2013vision} dataset and the subset of the DrivingStereo~\cite{drivingstereo} dataset. 
    All datasets are recorded in driving scene but the Synthia~\cite{ros2016synthia} is synthetic data, the KITTI and DrivingStereo are obtained from real world.
    The Synthia~\cite{ros2016synthia} dataset contains 50 sequences which have different combination of weathers, seasons and locations.
    To set similar disparity ranges in training and test, we resized Synthia~\cite{ros2016synthia} dataset images to half resolution as in \cite{tonioni2019learning}.
    We exploit stereo images from front direction only and there are 45,591 total number of stereo frames.
    The KITTI~\cite{geiger2013vision} dataset consists of 71 sequences and total 42,917 frames of stereo images and sparse depth maps. 
    Different from the scenario on the KITTI~\cite{geiger2013vision} dataset, we present an additional adaptation scenario that the models adapt to various unseen weather conditions using the subset of the DrivingStereo~\cite{drivingstereo} dataset.
    The DrivingStereo contains four different weather sequences (\ie cloudy, foggy, rainy, and sunny) that each sequence includes 500 stereo images with high quality labels obtained from multi-frame LiDAR points.

    \vspace{-10pt}
    \subsubsection{Metrics.}
    We evaluate the performance using two popular evaluation metrics, the percentage of pixels with disparity outliers larger than 3 (D1-all) and average end point error (EPE).
    Following the scheme of \cite{tonioni2019real,tonioni2019learning}, we average each score from all the frames which belong to the same sequence and reset the model to the base parameters at the next sequence, based on the definition of a sequence for different evaluation protocols.

    \vspace{-10pt}
    \subsubsection{Evaluation protocols.}
    We perform online stereo adaptation under three different settings according to the definition of the sequence, including short-, mid-, and long-term adaptation.
    For \textbf{short-term adaptation}, each sequence is defined as a distinct sequence provided by the dataset {\small (\eg \textit{2011\_09\_30\_drive\_0028\_sync})}.
    This setting is appeared in \cite{tonioni2019learning}.
    The sequences in \textbf{mid-term adaptation} are divided according to the environment (\ie \textit{City}, \textit{Residential}, \textit{Campus}, \textit{Road}).
    In \textbf{long-term adaptation}, we perform adaptation for all frames by concatenating all mid-term sequences.
    The mid- and long-term adaptation settings are shown in \cite{tonioni2019real} as short- and long-term adaptation.
    The implementation details are provided in supplementary material.

    \begin{table*}[t]
    \centering
    \caption{\textbf{Mid-term adaptation}: Performance comparison for several methods on the KITTI~\cite{geiger2013vision} dataset.
    }\label{tab:mid-range}\vspace{-5pt}
    \resizebox{\linewidth}{!}{
        \begin{tabular}{@{\extracolsep{4pt}}lcccccccccccc@{}}
        \hline
        \multirow{2}{*}{Method} & \multirow{2}{*}{Training} & \multirow{2}{*}{Adapt.}  
        & \multicolumn{2}{c}{City}  & \multicolumn{2}{c}{Residential}             
        & \multicolumn{2}{c}{Campus} & \multicolumn{2}{c}{Road}             
        & \multicolumn{2}{c}{Avg.}  
        \\ \cline{4-5}  \cline{6-7} \cline{8-9} \cline{10-11} \cline{12-13}
        &  &
        & \multicolumn{1}{l}{D1-all } & \multicolumn{1}{l}{EPE} 
        & \multicolumn{1}{l}{D1-all } & \multicolumn{1}{l}{EPE} 
        & \multicolumn{1}{l}{D1-all } & \multicolumn{1}{l}{EPE} 
        & \multicolumn{1}{l}{D1-all } & \multicolumn{1}{l}{EPE} 
        & \multicolumn{1}{l}{D1-all } & \multicolumn{1}{l}{EPE} 
        \\ \hline \hline
        DispNetC-GT & KITTI     & No                      
        &1.94   &0.68   &2.43   &0.77   &5.43   &1.10   &1.67   &0.69   &2.87   &0.81 \\
        MADNet-GT   & KITTI     & No    
        &2.05   &0.65   &2.67   &0.82   &6.87   &1.24   &1.57   &0.66   &3.29   &0.84 \\ \hline
        L2A-Disp.    & Synthia         & No      
        &12.78  &1.67   &12.80  &1.72   &17.57  &2.06   &12.34  &1.59   &13.87  &1.76 \\
        MADNet      & Synthia   & No  
        &38.78  &8.36   &35.73  &7.89   &40.59  &7.68   &38.31  &8.77   &38.35   &8.18 \\
        Ours-Disp.          & Synthia   & No
        &9.98   &1.47   &10.99  &1.62   &17.01  &2.06   &7.98   &1.33   &11.49  &1.62 \\
        Ours-MAD.           & Synthia   & No
        &15.51  &1.82   &14.24  &1.78   &22.40  &3.04   &15.61  &1.84   &16.94  &2.12 \\    \hline
        L2A-Disp.         & Synthia   & Full                    
        &2.05   &0.78   &2.57   &0.86   & 4.43  &1.07   &1.63   &0.77   &2.67   &0.87 \\
        MADNet      & Synthia   & Full  
        &2.11   &0.81   &2.79   &0.90   &6.24  &1.41   &1.60   &0.72   &3.19   &0.96 \\
        Ours-Disp.          & Synthia   & Full
        &2.03   &0.99   &2.46   &0.83   &4.21   &\textbf{1.02}   &1.58   &0.74   &2.57   &0.90 \\
        Ours-MAD.           & Synthia   & Full  
        &\textbf{1.55}   &\textbf{0.72}   &\textbf{1.55}   &\textbf{0.70}   &\textbf{3.84}   &1.08   &\textbf{1.15}   &\textbf{0.67}   &\textbf{2.02}   &\textbf{0.79} \\    \hline
        MADNet      & Synthia   & MAD                     
        &2.36   &0.84   &1.94   &0.77   &10.03  &1.70   &2.27   &0.83   &4.15   &1.04 \\
        MADNet      & Synthia   & MAD++                     
        &1.95   &0.80   &1.86   &0.76   &8.57  &1.65   &1.94   &0.80   &3.56   &0.99 \\
        Ours-MAD.           & Synthia   & MAD
        &1.63   &0.74   &1.62   &0.73   &4.16   &1.12   &1.23   &0.69   &2.16   &0.82 \\    \hlinewd{0.8pt}
    \end{tabular}\vspace{-20pt}
    }
    \end{table*}

\vspace{-10pt}
\subsection{Synthetic to Real Adaptation}
    \vspace{-5pt}
    We evaluate the performance corresponding to the different adaptation methods:
    \textit{No adaptation (No)}, which measures the performance for all sequences without performing adaptation from the base parameters to estimate the capacity of the initial parameters;
    \textit{Full adaptation (Full)}, which updates parameters of whole network;
    \textit{MAD adaptation (MAD)}, which performs faster modular adaptation on a prediction of certain resolution selected at every iteration by their own handcrafted method as proposed in \cite{tonioni2019real};
    \textit{MAD++ adaptation (MAD++)} is an extension from MAD and utilizes predictions obtained by handcrafted methods (\eg SGM~\cite{sgm}, WILD~\cite{tosi2017learning}) as proxy supervision.
    The cases of \textit{-GT} are regarded as supervised learning that is fine-tuned on KITTI 2012~\cite{geiger2012we} and 2015~\cite{menze2015object} datasets. In the experiment, (L2A, Ours)-Disp. and Ours-MAD. employ DispNetC and MADNet as base networks, respectively.

    \begin{table}[t]
    \centering
    \caption{\textbf{Short-term and Long-term adaptation}: Performance comparison for several methods on the KITTI~\cite{geiger2013vision} dataset.
    }\label{tab:short-range}\vspace{-5pt}
    \resizebox{0.67\linewidth}{!}
    {
        \begin{tabular}{@{\extracolsep{4pt}}lcccccc@{}}
        \hlinewd{0.8pt}
        \multirow{2}{*}{Method} & \multirow{2}{*}{Training} & \multirow{2}{*}{Adapt.} 
        & \multicolumn{2}{c}{Short-term} & \multicolumn{2}{c}{Long-term} 
        \\ \cline{4-5}  \cline{6-7} & & 
        & \multicolumn{1}{l}{D1-all } & \multicolumn{1}{l}{EPE}
        & \multicolumn{1}{l}{D1-all } & \multicolumn{1}{l}{EPE}
        \\ \hline \hline
        DispNetC-GT & KITTI     &No     &2.38   &0.77   &2.32   &0.75   \tabularnewline
        MADNet-GT   & KITTI     &No     &2.57   &0.75   &2.52   &0.78   \tabularnewline \hline
        
        L2A-Disp.   & Synthia   &No     &12.99  &1.73   &12.86  &1.70   \tabularnewline
        MADNet      & Synthia   &No     &27.63  &3.59   &36.77  &8.09   \tabularnewline
        Ours-Disp.  & Synthia   &No     &9.47   &1.21   &10.56  &1.57   \tabularnewline
        Ours-MAD.   & Synthia   &No     &11.59  &1.67   &23.50  &3.23   \tabularnewline \hline
        
        L2A-Disp.   & Synthia   &Full   &2.64   &0.84   &2.37   &0.84   \tabularnewline
        MADNet      & Synthia   &Full   &6.68   &1.31   &2.86   &0.93   \tabularnewline
        Ours-Disp.  & Synthia   &Full   &2.62   &0.81   &2.30   &0.82   \tabularnewline
        Ours-MAD.   & Synthia   &Full   &\textbf{2.00}  &\textbf{0.74}  &1.56   &0.71   \tabularnewline \hline
        
        MADNet      & Synthia   &MAD    &11.82  &1.90   &1.92   &0.75   \tabularnewline
        MADNet      & Synthia   &MAD++  &9.56   &1.61   &1.70   &0.75   \tabularnewline
        Ours-MAD.   & Synthia   &MAD    &2.64   &0.87   &\textbf{1.47}  &\textbf{0.70}  \tabularnewline   

        \hlinewd{0.8pt}
        \end{tabular}
    }
    \end{table}\vspace{-10pt}

    
    \vspace{-10pt}\subsubsection{Mid-term adaptation.}
    \tabref{tab:mid-range} shows the results according to the adaptation methods under mid-term adaptation setting.
    From the performance with \textit{No adaptation} (row 3-6), we observe that our method helps to learn better base parameters compared to other methods using the same base network. 
    Especially before the adaptation, with MADNet (row 4 and 6), our \model~significantly improves the performance of the base network by 20.19\% and 4.39 in terms of D1-all and EPE respectively, that demonstrates \model~is effective in a generalization capability.
    
    The results of \textit{Full adaptation} (row 7-10) and \textit{MAD adaptation} (row 11-13) show the adaptation capability of the base network.
    Our \model~outperforms previous methods with large margin in both metrics, achieving state-of-the-art performance on all domains.
    The \model~with MAD adaptation (row 13) outperforms MADNet with MAD++ adaptation that leverages the additional supervision and even L2A and MADNet with full adaptation (row 7 and 8) while enabling fast inference by adapting only a few parameters.
    As pointed out in \cite{tonioni2019real}, all adapted models perform worse on \textit{Campus} domain that has a small number of frames (1149) compared to the other domain (5674, 28067, 8027).
    The results on \textit{Campus} domain show that \model~adapt better than previous works with a small number of frames in all adaptation methods.\vspace{-5pt}

    \vspace{-10pt}
    \subsubsection{Short-term adaptation.}
        In~\tabref{tab:short-range}, to examine the adaptability in short sequences, we evaluate models on each sequence independently as represented in \cite{tonioni2019learning}. For each sequence, parameters are initialized at every beginning of sequences. Measured performance is first averaged in each sequence and then they are averaged out. 
        Thanks to fast adaptation speed and inherent robustness of our framework, we surpass the performance than previous works with a large margin.
        Especially, due to its light weight structure, inherent weakness of MADNet is maximized in short-term environment (row 4 and 8) because they requires a number of frames to be adapted. Nevertheless, MADNet with our framework shows superior results. This suggests that our framework is worthy to be developed with light weight networks.

 \begin{table*}[t]
    \centering
    \caption{\textbf{Short-term adaptation}: Performance comparison on the subset of DrivingStereo~\cite{drivingstereo} under different weather conditions.
    }\label{tab:other}\vspace{-5pt}
    \resizebox{\linewidth}{!}{
        \begin{tabular}{@{\extracolsep{4pt}}lcccccccccccc@{}}
        \hline
        \multirow{2}{*}{Method} & \multirow{2}{*}{Adapt.}  
        & \multicolumn{2}{c}{cloudy}  & \multicolumn{2}{c}{foggy}             
        & \multicolumn{2}{c}{rainy} & \multicolumn{2}{c}{sunny}
        & \multicolumn{2}{c}{Avg.} 
        \\ \cline{3-4}  \cline{5-6} \cline{7-8} \cline{9-10} \cline{11-12} 
        & 
        & \multicolumn{1}{l}{D1-all } & \multicolumn{1}{l}{EPE} 
        & \multicolumn{1}{l}{D1-all } & \multicolumn{1}{l}{EPE} 
        & \multicolumn{1}{l}{D1-all } & \multicolumn{1}{l}{EPE} 
        & \multicolumn{1}{l}{D1-all } & \multicolumn{1}{l}{EPE} 
        & \multicolumn{1}{l}{D1-all } & \multicolumn{1}{l}{EPE} 
        \\ \hline \hline
        MADNet      & No                     
        &56.83	&19.16	&70.14	&23.85	&54.20	&19.20	&51.30	&16.08   &58.12  &19.57   \\
        Ours-MAD.   & No
        &32.76	&5.34	&37.25	&6.55	&34.06	&4.78	&30.00	&4.58   &33.52  &5.31  
        \\    \hline
        MADNet      & Full  
        &15.71	&3.11	&18.09	&3.28	&18.37	&2.86	&14.71	&2.63   &16.72  &2.97     
        \\
        Ours-MAD.   & Full  
        &\textbf{7.00}	&\textbf{1.28}	&\textbf{7.25}	&\textbf{1.39}	&15.31	&2.52	&\textbf{8.33}	&\textbf{1.49}   &\textbf{9.47}   &1.67  
        \\    \hline
        MADNet      & MAD                     
        &28.46	&6.65	&33.56	&6.13	&31.34	&5.79	&27.10	&6.15   &30.11  &6.18  
        \\
        Ours-MAD.   & MAD
        &8.46	&1.46	&8.57	&1.46	&\textbf{11.99}	&\textbf{1.69}	&8.90	&1.52   &9.48   &\textbf{1.53} 
        \\    \hlinewd{0.8pt}
    \end{tabular}
    }
    \end{table*}

        We conduct additional experiments on the DrivingStereo~\cite{drivingstereo} dataset to hypothesize more difficult scenarios under various weather conditions.
        Specifically, we train all models on Synthia~\cite{ros2016synthia} dataset and evaluate the performance on each sequence including four types of weather conditions in the short-term adaptation setting.
        As shown in \tabref{tab:other}, our model outperforms MADNet~\cite{tonioni2019real} with a large margin for all novel weather conditions.
        In particular, our method represents error rates of about half those of MADNet~\cite{tonioni2019real}.
        The implementation details and additional experimental results are provided in supplementary material.\vspace{-5pt}

    \vspace{-10pt}
    \subsubsection{Long-term adaptation.}
    The adaptation on a long sequence followed by various environments without network resets can be regarded as the most practical scenario in the real world.
    To simulate this scenario, we report the results evaluated on the concatenation of four environments of the KITTI~\cite{geiger2013vision} datase ($\sim$ 43000 frames) in \tabref{tab:short-range}.
    As analyzed in \cite{tonioni2019real}, the results show much smaller average errors than the mid- and short-term adaptation for all adaptation methods, as the length of the sequence increased.
    Among them, our \model~shows drastically improved performance and significantly outperforms previous works.
    Therefore, \model~framework can be further improved, continually adapting to the real world environment.


    \vspace{-10pt}\subsection{Analysis}\vspace{-3pt}
    \begin{wraptable}{r}{6.3cm}\vspace{-33pt}
    \caption{Ablation studies for \model Net and the meta-learning framework (ML) evaluated on the KITTI~\cite{geiger2013vision} dataset under short-term adaptation setting.}\label{tab:ablation}
    \begin{tabular}{
    >{\centering}m{0.31\linewidth}>{\centering}m{0.13\linewidth}
    >{\centering}m{0.18\linewidth}>{\centering}m{0.18\linewidth}>{\centering}m{0.13\linewidth}}
        \hlinewd{0.8pt}
        \model Net  &   ML   &   Adapt.  &   D1-all  &   EPE \tabularnewline
        \hline  \hline
        \xmark  &   \xmark  &   Full    &   6.68    &   1.31    \tabularnewline
        \cmark  &   \xmark  &   Full    &   8.06    &   1.47    \tabularnewline
        \xmark  &   \cmark  &   Full    &   3.12    &   0.96    \tabularnewline
        \cmark  &   \cmark  &   Full    &   \textbf{2.00}    &   \textbf{0.74}    \tabularnewline
        \hlinewd{0.8pt}
        
    \end{tabular}\vspace{-20pt}
    \end{wraptable}
    
    \subsubsection{Ablation study.}
    To investigate the effectiveness of the components of within our model, we conduct ablation experiments on the KITTI~\cite{geiger2013vision} dataset according to \model Net and the meta-learning framework (ML), as shown in \tabref{tab:ablation}.
    Note that we use MADNet as the base network and evaluate the performance using the full adaptation under the short-term adaptation setting for all experiments in this section.
    As a baseline, we remove all components of the proposed method such that the first row in \tabref{tab:ablation} corresponds to MADNet~\cite{tonioni2019real}. \vspace{-5pt}
    
    \paragraph{Effectiveness of \model Net.}
    To validate the effectiveness of the point-wise backpropagation, we ablate \model Net and apply $\ell_1$ loss between the initial disparity $\hat{d}$ and groundtruth $d$ instead of the point loss in Alg. 1.
    The comparison between the third row of \tabref{tab:ablation} and the full use of components shows \model Net contributes $1.12\%$ and $0.22$ in terms of D1-all error and EPE and demonstrates fixing local detriments is simple yet effective to improve the robustness of the stereo model.

    \paragraph{Effectiveness of ML.}
    As described in \ref{l2f}, the performance improvement of the base network using the point loss is not guaranteed without meta-learning.
    The results in the second row of \tabref{tab:ablation} show poor performance even than the baseline.
    The comparison between the first and third rows of \tabref{tab:ablation} further validates the effectiveness of the ML framework, showing significant performance improvements.
    Finally, the state-of-the-art performance is shown by demonstrating the advantage of the full use of all components of learning to fix the base network through meta-learning. \vspace{-5pt}
    
    \begin{figure*}[t]
    \begin{center}
       \includegraphics[width=1\linewidth]{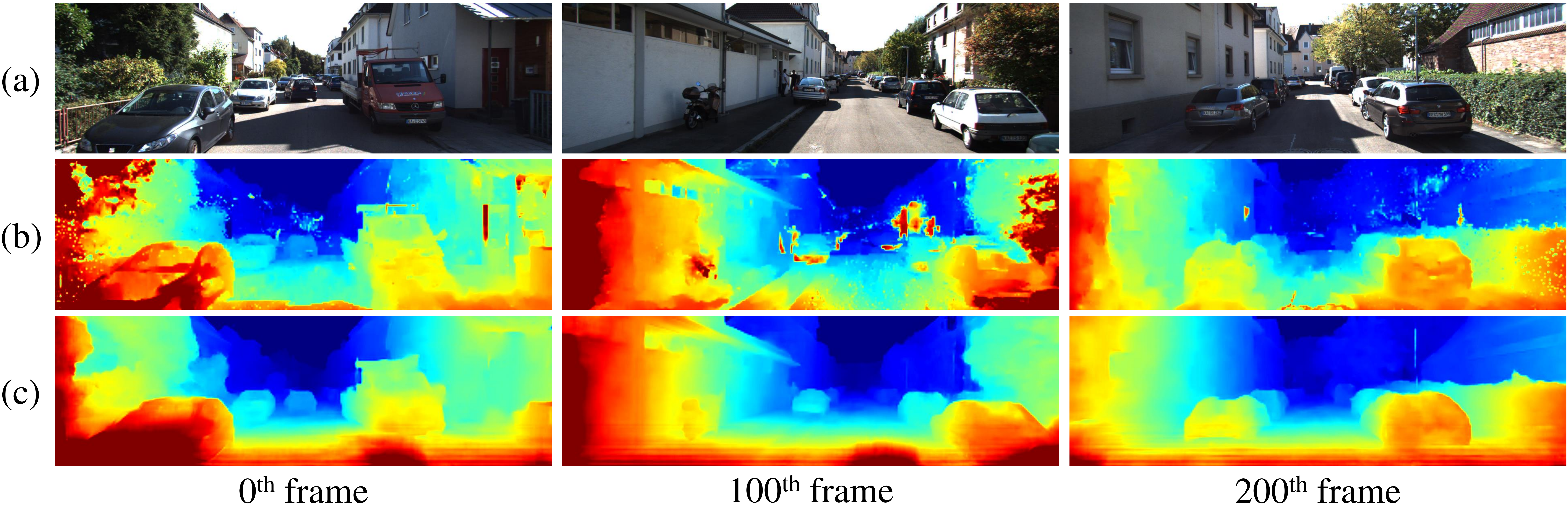}
    \end{center}\vspace{-17pt}
       \caption{Disparity maps predicted using MADNet as the base network on the KITTI~\cite{geiger2013vision} sequence. (a) Left images, (b) MADNet with MAD adaptation~\cite{tonioni2019real}, (c) Ours-MAD. with MAD adaptation. Red pixel values indicate closer objects.}
    \label{fig:convergence1}
    \end{figure*}
    
    \begin{wrapfigure}{r}{6cm}
    \vspace{-35pt}
    \begin{center}
    \includegraphics[width=1\linewidth]{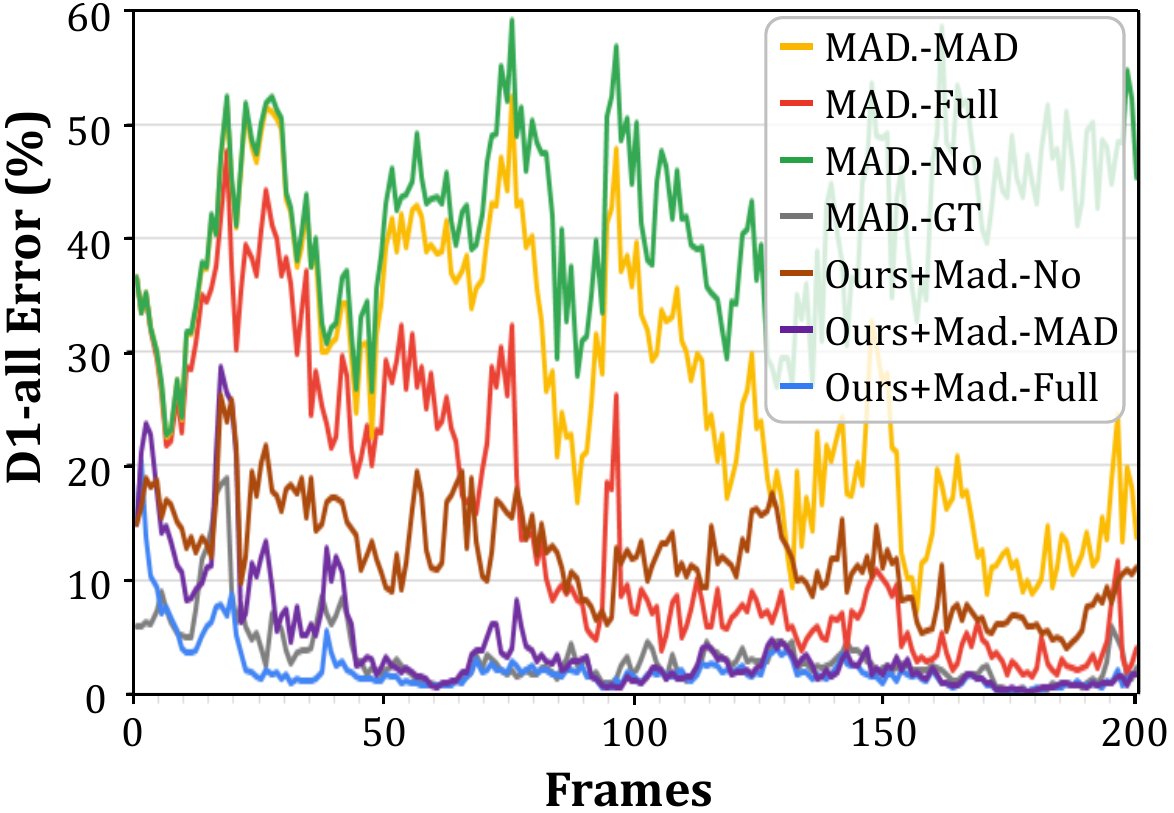}
    \end{center}
    \vspace{-20pt}
       \caption{D1-all error (\%) across frames in sequence from the KITTI~\cite{geiger2013vision} dataset with respect to the adaptation methods.}\vspace{-20pt}
    \label{fig:convergence2}
    \end{wrapfigure}
    \vspace{-5pt}\subsubsection{Convergence.}
    In \figref{fig:convergence1}, we evaluate the qualitative results related to the convergence analysis.
    The results contain good initialized parameters (first column), fast adaptation (second column), and convergence to low errors (last column) as analyzed above.
    As the adaptation proceeds, the MADNet (row 2) estimates better prediction, yet still shows a high error while Ours-MAD (row 3) shows not only robust initial performance but also faster convergence to low errors.
    
    To analyze and compare the adaptation cost corresponding to the methods, we visualize the adaptation performance over frames of the sequence from the KITTI~\cite{geiger2013vision} dataset in \figref{fig:convergence2}\footnote{
    The results with DispNetC are shown in supplementary materials.}.
    In overall view, the results show that our \model~adapts faster than \cite{tonioni2019real} and converges with lower errors regardless of the adaptation method. Furthermore, the comparison between \textbf{MAD.-No} (green) and \textbf{Ours+Mad.-No} (brown) shows the effectiveness of the initial base parameters.
    The comparison between \textbf{MAD.-MAD} (yellow) and \textbf{Ours+Mad.-MAD} (purple) shows that the performance is improved by \model~with MAD adaptation, while \textbf{MAD.-Full} (red) overtakes from the about $100$-th frame.
    Finally, \textbf{Ours+Mad.-Full} (blue) adapts faster than all the other methods and converges to the low D1-all error, showing comparable performance with \textbf{MAD.-GT} (gray) fine-tuned with ground-truth.
    The additional experimental results and analysis are shown in supplementary material.

    \vspace{-5pt}\subsubsection{Comparison with domain generalization methods.}
    To argue the practicality of online stereo adaptation, we compare our model with the state-of-the-art domain generalization (DG) methods~\cite{dsmnet,cfnet,lipson2021raft} that the stereo models are trained on the synthetic dataset and evaluated on the unseen real dataset without the additional adaptation.\\ \vspace{-10pt}
    \begin{wraptable}{r}{6.3cm}\vspace{-45pt}
    \caption{Comparisons of domain generalization methods and our method evaluated on the KITTI~\cite{geiger2013vision} dataset.}\label{tab:generalization}
    \begin{tabular}{
    >{\raggedright}m{0.31\linewidth}>{\centering}m{0.18\linewidth}
    >{\centering}m{0.17\linewidth}>{\centering}m{0.13\linewidth}>{\centering}m{0.13\linewidth}}
         \hlinewd{0.8pt}
         Method &   Adapt.  &   D1-all  &   EPE &   FPS \tabularnewline
         \hline \hline
         DSMNet~\cite{dsmnet} & N/A &1.59  &0.68   &1.30   \tabularnewline
         CFNet~\cite{cfnet}  & N/A &  1.93    &   0.97    &   4.27    \tabularnewline
         Raft~\cite{lipson2021raft}  & N/A &  1.66    &   0.71    &   22.44    \tabularnewline
         \hline
         MADNet~\cite{tonioni2019real}  &   MAD     &   1.95    &   0.82    &   35.7    \tabularnewline
         Ours-MAD.  &  MAD &    1.47    &   0.70 &   35.7   \tabularnewline
         \hlinewd{0.8pt}
    \end{tabular}\vspace{-20pt}
    \end{wraptable}
    For a fair comparison, the models are pretrained on the SceneFlow~\cite{dispnet} dataset and evaluated on the KITTI~\cite{geiger2013vision} dataset\footnote{We conducted a custom experiment using a publicly available code for each paper.}.
    Note that we report the performance of our model measured under the long-term adaptation setting.
    As shown in \tabref{tab:generalization}\footnote{For Raft-stereo~\cite{lipson2021raft}, a real-time version was employed which shows a much faster inference speed.}, our model not only outperforms the generalization approaches in terms of D1-all error but also shows about $\times27$, $\times8.3$, and $\times1.6$ faster inference speed than \cite{dsmnet}, \cite{cfnet}, and \cite{lipson2021raft}, respectively, despite of additional adaptation steps.
    While the domain generalization approaches~\cite{dsmnet,cfnet,lipson2021raft} estimate the depth maps without the adaptation, they require a large number of parameters to obtain a generalized stereo model, making them impractical.
    This is worth noting that our method has high applicability to the practical application such as autonomous driving in terms of accuracy and inference speed.
    
\section{Conclusion}
\vspace{-7pt}
In this paper, we proposed \model, a novel meta-learning framework to effectively adapt any deep stereo models in online setting.
Compared with previous online stereo adaptation approaches facing global domain bias problem to synthetic data, our model can induce maximal performance of the base stereo networks by the proposed the auxiliary network \model Net and learning-to-fix strategy, that can adapt well to the fine-grained domain gap.
Our extensive experiments show \model~achieves state-of-the-art results, outperforming several online stereo adaptation methods in a wide variety of environments.
In addition, the results demonstrate that \model~is capable of improving the generalization ability of the stereo models.
\vspace{-7pt}

\clearpage
%
%
\bibliographystyle{splncs04}
\bibliography{egbib}

\pagestyle{headings}
\def\ECCVSubNumber{2217}  

\title{PointFix: Learning to Fix Domain Bias for Robust Online Stereo Adaptation \\ 
- Supplementary Materials -} 

\titlerunning{Learning to Fix Domain Bias for Robust Online Stereo Adaptation}
%
\author{Kwonyoung Kim$^{1}$ \quad
Jungin Park$^{1}$ \quad
Jiyoung Lee$^{2}$ \\
Dongbo Min$^{3}$ \quad
Kwanghoon Sohn$^{1}$\thanks{Corresponding author.}}

\institute{
$^{1}$Yonsei University \quad $^{2}$NAVER AI Lab \quad $^{3}$Ewha Womans University \\ \vspace{3pt}
{\tt\small $\lbrace$kyk12, newrun, khsohn$\rbrace$@yonsei.ac.kr} \quad \\
{\tt\small lee.j@navercorp.com} \quad
{\tt\small dbmin@ewha.ac.kr}}


%
%
%
\maketitle

    In this supplement, we provide the implementation details and the additional experimental results.
    First, we provide the training and adaptation details used in our experiments, and the network configuration of \model Net.
    To demonstrate the effectiveness of \model, we extend the experiments, including the convergence analysis, synthetic to synthetic scenarios, qualitative results and additional ablation studies, in the following section.
    The videos for the predicted disparity sequence (\textit{2011\_09\_30\_drive\_0027\_sync} of the KITTI~\cite{geiger2013vision} dataset) and the 3D reconstruction from the predicted disparity are provided in .mp4 format. Please refer our video supplementary files (`Sim-to-Real.zip').

\section{Implementation Details}
\subsubsection{Additional datasets.}
    In this section, we introduce datasets used to conduct the additional experiments, including comparisons with domain generalization approaches in the main paper and \textit{synthetic to synthetic adaptation} in this document.
    Specifically, we use the SceneFlow~\cite{dispnet} dataset to train the models evaluated on comparisons with domain generalization methods~\cite{dsmnet,cfnet}.
    The SceneFlow is a synthetic dataset and contains 39,000 stereo frames with 960 $\times$ 540 pixel resolution.
    We use the FlyingThings3D (F3D)~\cite{dispnet} dataset, which is one of the subsets of the SceneFlow dataset, only to train the models for the experiments in this supplement.
    In addition, we use the Virtual KITTI 2~\cite{vkitti2} dataset as a test benchmark in \secref{sec:32}, that contains photo-realistic and synthetic stereo images recreated from the KITTI~\cite{geiger2012we} dataset.
    This dataset consists of 5 separate scenes corresponding to different locations and each scene is transformed to represent 6 different weather and lighting conditions (\ie clone, fog, morning, overcast, rain, and sunset) for the same scene.
    Therefore, we use a total of 30 sequences for short-, mid-, and long-term adaptation. Each sequence consists of at least 233 frames and frame resolution is 1242 $\times$ 375.
    
\subsubsection{Training.}
    In our experiments, we use TensorFlow library~\cite{tensorflow2015-whitepaper} and a single NVIDIA RTX A6000 48GB for training and NVIDIA TITAN RTX 24GB for inference.
    For the short-term experiment in the main paper, the test images of DrivingStereo~\cite{drivingstereo} dataset are rescaled into 703$\times$320 for disparity range.
    For such a reason, we resize whole images in Synthia~\cite{ros2016synthia} dataset to half resolution as in \cite{tonioni2019learning}.
    All of initial parameters of base models are pretrained on the F3D dataset which is a synthetic dataset provided from \cite{dispnet}.
    Unless specified, we used the Synthia~\cite{ros2016synthia} dataset to train the base model with our method. 
    For our model that uses the DispNetC~\cite{dispnet} as the base network, we set the inner loop learning rates $\alpha$ and the outer loop learning rate $\beta$ to $10^{-4}$ during the first 40k iterations, and $\alpha=10^{-5}$ and $\beta=10^{-4}$ for the last 10k iterations.
    For the model that uses the MADNet~\cite{tonioni2019real} as the base network, we use $\alpha=10^{-5}$ and $\beta=10^{-4}$ during 30k iterations, respectively.

\subsubsection{Online adaptation.}
    At inference, the parameters of the base network are continuously updated with the reconstruction loss.
    For models trained on Synthia~\cite{ros2016synthia}, we set the adaptation learning rate to $10^{-4}$ for their best performance. 
    For our model with the MADNet as the base network, the learning rate for short-term adaptation is set to $10^{-4}$ and $10^{-5}$ is given for the mid-term and long-term \textit{Full adaptation} experiments.
    For models trained on the F3D~\cite{dispnet} or SceneFlow~\cite{dispnet}, we set the adaptation learning rate to $10^{-5}$ and $10^{-4}$ for the \textit{Full adaptation} and \textit{MAD adaptation}, respectively.
    
    We compute the reconstruction loss differently according to the base network and the adaptation method due to the different structures of the network and the adaptation strategy.
    Specifically, the reconstruction loss for the DispNet~\cite{dispnet} is computed on the final predicted disparity map following \cite{tonioni2019learning}.
    The MADNet~\cite{tonioni2019real} with \textit{Full adaptation} computes the reconstruction loss for all disparity maps at every single module, while a single disparity map is chosen for the reconstruction loss with the MAD adaptation by the strategy proposed in \cite{tonioni2019real}.

\subsubsection{Network architecture.}
    As described in the main paper, our \model Net consists of two main modules: a feature extraction module and a point-wise prediction module.
    We depict the network configuration using the DispNetC~\cite{dispnet} and the MADNet~\cite{tonioni2019real} as the base network in \tabref{tab:configuration_disp} and \tabref{tab:configuration_mad}, respectively.
    The feature extraction module consists of three convolutional layers and the point-wise prediction module contains four fully-connected (FC) layers.
    Such lightweight model-agnostic architecture enables \model Net to be applied regardless of the base network and makes the base network maintain inference speed.
        
    \clearpage
    \begin{table}[!h]
    	\begin{center}
    	\caption{Network configuration using DispNetC~\cite{dispnet} as the base network. }
    	\label{tab:configuration_disp}
    	\resizebox{0.6\linewidth}{!}{
    		\begin{tabular}{l|ccc}
    			
    			\hlinewd{0.8pt}
    			\multicolumn{4}{c}{DispNetC} \tabularnewline
    			\hlinewd{0.8pt}
    			Layer & Ch I/O & Input & Output \tabularnewline
    			\hline
    			\hline
    			conv1 & 3~/~64 & $I^l, I^r$ & $\text{feat1}^{l,r}$  \tabularnewline
    			conv2 & 64~/~128 &  $\text{feat1}^{l,r}$ &  $\mathbf{f}^{l,r}$                  \tabularnewline
    			conv\_redir &  128~/~64   &   $\mathbf{f}^l$    & $\text{feat\_redir}$      \tabularnewline
    			corr &  128~/~81   &   $\mathbf{f}^{l,r}$    & $\mathbf{c}$      \tabularnewline
    			conv\_3a &  145~/~256   &   $\mathbf{c}, \text{feat\_redir}$    & $\text{feat3a}$      \tabularnewline
    			conv\_3b &  256~/~256   &   $\text{feat3a}$    & $\text{feat3b}$      \tabularnewline
    			conv\_4a &  256~/~512   &   $\text{feat3b}$    & $\text{feat4a}$      \tabularnewline
    			conv\_4b &  512~/~512   &   $\text{feat4a}$    & $\text{feat4b}$      \tabularnewline
    			conv\_5a &  512~/~512   &   $\text{feat4b}$    & $\text{feat5a}$      \tabularnewline
    			conv\_5b &  512~/~512   &   $\text{feat5a}$    & $\text{feat5b}$      \tabularnewline
    			conv\_6a &  512~/~1024   &   $\text{feat5b}$    & $\text{feat6a}$      \tabularnewline
    			conv\_6b &  1024~/~1024   &   $\text{feat6a}$    & $\text{feat6b}$      \tabularnewline
    			
    			pr6+loss6 &  1024~/~1   &   $\text{ifeat6}$    & $\text{pr6}$      \tabularnewline
    			upconv5 &  1024~/~512   &   $\text{feat6b}$    & $\text{upfeat5}$      \tabularnewline
    			iconv5 &  1024~/~512   &   $\text{upfeat5,pr6,feat5b}$    & $\text{ifeat5}$      \tabularnewline
    			
    			pr5+loss5 &  512~/~1   &   $\text{ifeat5}$    & $\text{pr5}$      \tabularnewline
    			upconv4 &  512~/~256   &   $\text{feat5b}$    & $\text{upfeat4}$      \tabularnewline
    			iconv4 &  769~/~256   &   $\text{upfeat4,pr5,feat4b}$    & $\text{ifeat4}$      \tabularnewline
    			
    			pr4+loss4 &  256~/~1   &   $\text{ifeat4}$    & $\text{pr4}$      \tabularnewline
    			upconv3 &  256~/~128   &   $\text{feat4b}$    & $\text{upfeat4}$      \tabularnewline
    			iconv3 &  385~/~128   &   $\text{upfeat3,pr4,feat3b}$    & $\text{ifeat3}$      \tabularnewline
    			
    			pr3+loss3 &  128~/~1   &   $\text{ifeat3}$    & $\text{pr3}$      \tabularnewline
    			upconv2 &  128~/~64   &   $\text{feat3b}$    & $\text{upfeat2}$      \tabularnewline
    			iconv2 &  193~/~64   &   $\text{upfeat2,pr3},\mathbf{f}^{l}$    & $\text{ifeat2}$      \tabularnewline
    			
    			pr2+loss2 &  64~/~1   &   $\text{ifeat2}$    & $\text{pr2}$      \tabularnewline
    			upconv1 &  64~/~32   &   $\text{feat2b}$    & $\text{upfeat1}$      \tabularnewline
    			iconv1 &  97~/~32   &   $\text{upfeat1,pr2,feat1b}$    & $\text{ifeat1}$      \tabularnewline
    			
    			pr1+loss1 &  32~/~1   &   $\text{ifeat1}$    & $\text{pr1}$      \tabularnewline

    			\hlinewd{0.8pt}
    			\multicolumn{4}{c}{\model Net: Feature Extraction Module} \tabularnewline
    			\hlinewd{0.8pt}
    			Layer & Ch I/O & Input & Output \tabularnewline
    			\hline
    			\hline
    			conv1 & 5~/~32  &   $I^l, \hat{d}, d$   & $\mathbf{z}^{1}$  \tabularnewline
    			conv2 & 32~/~64 &  $\mathbf{z}^{1}$ &  $\mathbf{z}^{2}$                  \tabularnewline
    			conv3 &  64~/~128   &   $\mathbf{z}^{2}$    & $\mathbf{z}^{c}$      \tabularnewline
    			\hlinewd{0.8pt}
    			\multicolumn{4}{c}{\model Net: Point-wise Prediction Module} \tabularnewline
    			\hlinewd{0.8pt}
    			Layer & Ch I/O & Input & Output \tabularnewline
    			\hline
    			\hline
    			concat. & 64, 81~/~145 & $\mathbf{c}, \mathbf{f}^{l}$ & $\mathbf{z}^{b}$ \tabularnewline
    			concat. & 145, 128~/~273 & $\mathbf{z}^{b}$, $\mathbf{z}^{c}$ & $\Pi(\mathbf{z}^{b}, \mathbf{z}^{c})$ \tabularnewline
    			FC1 & 273~/~273 & $\Pi(\mathbf{z}^{b}, \mathbf{z}^{c})_{ij}$ & $\mathbf{x}^{1}_{ij}$ \tabularnewline
    			FC2 & 273~/~273 & $\mathbf{x}^{1}_{ij}$ & $\mathbf{x}^{2}_{ij}$ \tabularnewline
    			FC3 & 273~/~273 & $\mathbf{x}^{2}_{ij}$ & $\mathbf{x}^{3}_{ij}$\tabularnewline
    			FC4 & 273~/~1 &  $\mathbf{x}^{3}_{ij}$ & $r_{ij}$\tabularnewline
    			\hlinewd{0.8pt}
    		\end{tabular}
    		}
    	\end{center}
    	
    \end{table}
    
    \clearpage
    \begin{table}[!h]
    	\begin{center}
    	\caption{Network configuration using MADNet~\cite{tonioni2019real} as the base network.
        }
    	\label{tab:configuration_mad}
    	\resizebox{0.5\linewidth}{!}{
    		\begin{tabular}{l|ccc}
    			
    			\hlinewd{0.8pt}
    			\multicolumn{4}{c}{MADNet: Feature Extractor} \tabularnewline
    			\hlinewd{0.8pt}
    			Layer & Ch I/O & Input & Output \tabularnewline
    			\hline
    			\hline
    			conv1 & 3~/~16 & $I^l, I^r$ & $\text{feat}_0$  \tabularnewline
    			conv2 & 16~/~16 &  $\text{feat1}$ &  $\text{feat}_1$                  \tabularnewline
    			conv3 &  16~/~32   &   $\text{feat}_1$    & $\text{feat}_{1\_1}$      \tabularnewline
    			conv4 &  32~/~32   &   $\text{feat}_{1\_1}$    & $\text{feat}_2 (\mathbf{f})$      \tabularnewline
    			conv5 &  32~/~64   &   $\text{feat}_2$    & $\text{feat}_{2\_1}$      \tabularnewline
    			conv6 &  64~/~64   &   $\text{feat}_{2\_1}$    & $\text{feat}_3$      \tabularnewline
    			conv7 &  64~/~96   &   $\text{feat}_3$    & $\text{feat}_{3\_1}$      \tabularnewline
    			conv8 &  96~/~96   &   $\text{feat}_{3\_1}$    & $\text{feat}_4$      \tabularnewline
    			conv9 &  96~/~128   &   $\text{feat}_4$    & $\text{feat}_{4\_1}$      \tabularnewline
    			conv10 &  128~/~128   &   $\text{feat}_{4\_1}$    & $\text{feat}_5$      \tabularnewline
    			conv11 &  128~/~192   &   $\text{feat}_5$    & $\text{feat}_{5\_1}$      \tabularnewline
    			conv12 &  192~/~192   &   $\text{feat}_{5\_1}$    & $\text{feat}_6$      \tabularnewline
    			
    			\hlinewd{0.8pt}
    			\multicolumn{4}{c}{MADNet: Stereo Estimation network} \tabularnewline
    			\hlinewd{0.8pt}
    			Layer & Ch I/O & Input & Output \tabularnewline
    			\hline
    			\hline
    			conv1 & C~/~128 & $D_{n+1}, \text{feat}_{n}^{l,r}$ &  $\text{feat}_{1}^{\text{SE}}$ \tabularnewline
    			conv2 & 128~/~128 &  $\text{feat}_{1}^{\text{SE}}$ &  $\text{feat}_{2}^{\text{SE}}$                  \tabularnewline
    			conv3 &  128~/~96   &   $\text{feat}_{2}^{\text{SE}}$    & $\text{feat}_{3}^{\text{SE}}$      \tabularnewline
    			conv4 &  96~/~64   &   $\text{feat}_{3}^{\text{SE}}$    & $\text{feat}_4^{\text{SE}}$     \tabularnewline
    			conv5 &  64~/~32   &   $\text{feat}_4^{\text{SE}}$    & $\text{feat}_5^{\text{SE}}$      \tabularnewline
    			conv6 &  32~/~1   &   $\text{feat}_5^{\text{SE}}$    & $D_{n}$      \tabularnewline
    			
    			\hlinewd{0.8pt}
    			\multicolumn{4}{c}{MADNet: Residual Refinement network} \tabularnewline
    			\hlinewd{0.8pt}
    			Layer & Ch I/O & Input & Output \tabularnewline
    			\hline
    			\hline
    			conv1 & C~/~128 & $\text{feat}_n^l,D_{n}^*$ &  $\text{feat}_1^{\text{R}}$ \tabularnewline
    			conv2 & 128~/~128 &  $\text{feat}_1^{\text{R}}$ &  $\text{feat}_2^{\text{R}}$                  \tabularnewline
    			conv3 &  128~/~128   &   $\text{feat}_2^{\text{R}}$    & $\text{feat}_3^{\text{R}}$      \tabularnewline
    			conv4 &  128~/~96   &   $\text{feat}_3^{\text{R}}$    & $\text{feat}_4^{\text{R}}$     \tabularnewline
    			conv4 &  96~/~64   &   $\text{feat}_4^{\text{R}}$    & $\text{feat}_5^{\text{R}}$     \tabularnewline
    			conv5 &  64~/~32   &   $\text{feat}_5^{\text{R}}$    & $\text{feat}_6^{\text{R}}$      \tabularnewline
    			conv6 &  32~/~1   &   $\text{feat}_6^{\text{R}}$    & $R_{n}$      \tabularnewline

    			\hlinewd{0.8pt}
    			\multicolumn{4}{c}{\model Net: Feature Extraction Module} \tabularnewline
    			\hlinewd{0.8pt}
    			Layer & Ch I/O & Input & Output \tabularnewline
    			\hline
    			\hline
    			conv1 & 5~/~32 & $I^l, \hat{d}, d$ & $\mathbf{z}^{1}$  \tabularnewline
    			conv2 & 32~/~64 &  $\mathbf{z}^{1}$ &  $\mathbf{z}^{2}$                  \tabularnewline
    			conv3 &  64~/~128   &   $\mathbf{z}^{2}$    & $\mathbf{z}^{c}$      \tabularnewline
    			\hlinewd{0.8pt}
    			\multicolumn{4}{c}{\model Net: Point-wise Prediction Module} \tabularnewline
    			\hlinewd{0.8pt}
    			Layer & Ch I/O & Input & Output \tabularnewline
    			\hline
    			\hline
    			corr. & 32, 32~/~81 & $\mathbf{f}^{l,r}$ & $\mathbf{c}$  \tabularnewline
    			concat. & 32, 81~/~113 & $\mathbf{c}, \mathbf{f}^{l}$ & $\mathbf{z}^{b}$ \tabularnewline
    			concat. & 113, 128~/~241 & $\mathbf{z}^{b}$, $\mathbf{z}^{c}$ & $\Pi(\mathbf{z}^{b}, \mathbf{z}^{c})$ \tabularnewline
    			FC1 & 241~/~241 & $\Pi(\mathbf{z}^{b}, \mathbf{z}^{c})_{ij}$ & $\mathbf{x}^{1}_{ij}$ \tabularnewline
    			FC2 & 241~/~241 & $\mathbf{x}^{1}_{ij}$ & $\mathbf{x}^{2}_{ij}$ \tabularnewline
    			FC3 & 241~/~241 & $\mathbf{x}^{2}_{ij}$ & $\mathbf{x}^{3}_{ij}$\tabularnewline
    			FC4 & 241~/~1 & $\mathbf{x}^{3}_{ij}$ & $r_{ij}$\tabularnewline
    			\hlinewd{0.8pt}
    		\end{tabular}
    		}
    	\end{center}
    	
    \end{table}

\section{Additional Results}
    To show the superiority of our model, we further provide  experimental results.
    In \secref{sec:31}, we first analyze the performance convergence of our \model~and previous works~\cite{tonioni2019learning,tonioni2019real} by expanding experiments shown in Sec. 5.2 of the main paper.
    We also evaluate the online adaptation performance on the synthetic-to-synthetic adaptation setting using the F3D~\cite{dispnet}, Virtual KITTI~\cite{vkitti2} and Synthia~\cite{ros2016synthia} datasets in \secref{sec:32}.
    Finally, we provide additional qualitative results in \secref{sec:33}, including the qualitative comparisons with the online adaptation methods~\cite{tonioni2019learning,tonioni2019real} and domain generalization methods~\cite{dsmnet,cfnet} on the KITTI~\cite{geiger2013vision} dataset.

\subsection{Convergence Analysis}\label{sec:31}
\subsubsection{Short-term adaptation.}
    To prove the incomparable stability of our method, we display the adaptation performance over frames as contrasted with the previous methods~\cite{tonioni2019real} that stand on the DispNetC~\cite{dispnet} in a sequence\footnote{`\textit{2011\_09\_26\_drive\_101\_sync}' sequence is used} from the KITTI~\cite{geiger2013vision} in \figref{fig:1}.
    While \textbf{Ours+Disp.-Full} (blue) is slightly better than \textbf{L2A+Disp.-Full} (red) using full sequence for adaptation, the superior quality of the initial parameters of the base network using \model~is presented in the comparison between \textbf{L2A+Disp.-No} (yellow) and \textbf{Ours+Disp.-No} (green). The result shows \model~is more robust against new environmental changes, achieving better performance without the adaptation.
    Moreover, the method with full adaptation shows almost similar performance to \textbf{Disp.-GT} (gray) that is fine-tuned using the KITTI~\cite{geiger2012we,menze2015object} training sets.
    Note that the results using MADNet as the base network are shown in Fig. 4 of the main paper.

    \begin{figure*}
       \centering
    \subfigure{
    \begin{minipage}{0.5\linewidth}\centering
    \includegraphics[width=\columnwidth]{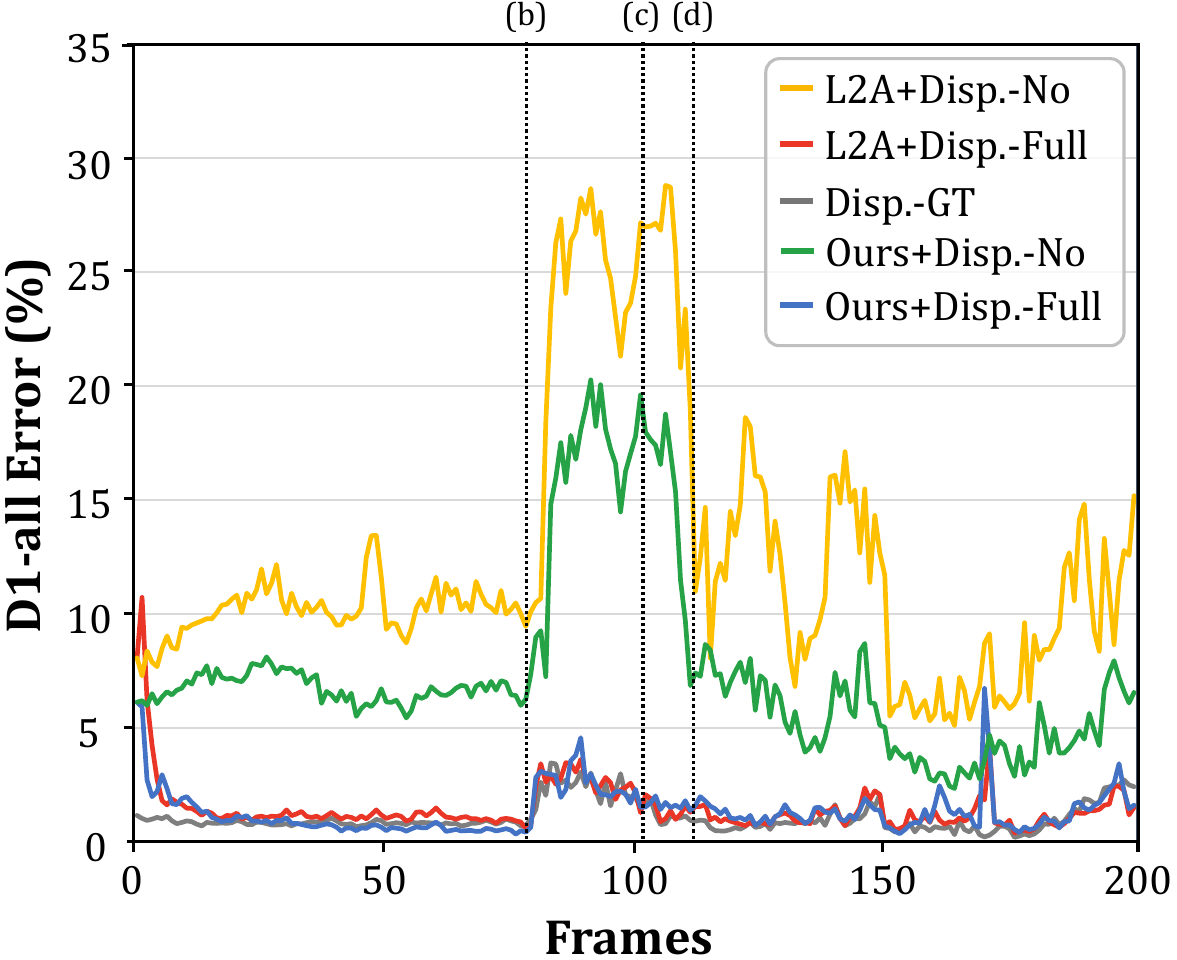}\hfill
    {(a) D1-all error (\%) across frames}\\
    \end{minipage}\hfill
    }\hfill
    \subfigure{
    \begin{minipage}{0.45\linewidth}\centering
    \includegraphics[width=\columnwidth]{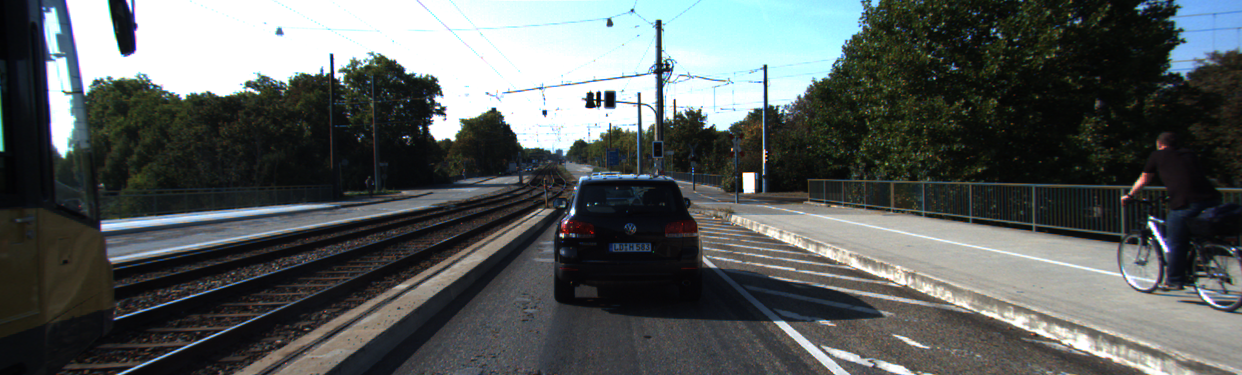}\hfill
    {(b) Input left image at 78th frame}\\
    \includegraphics[width=\columnwidth]{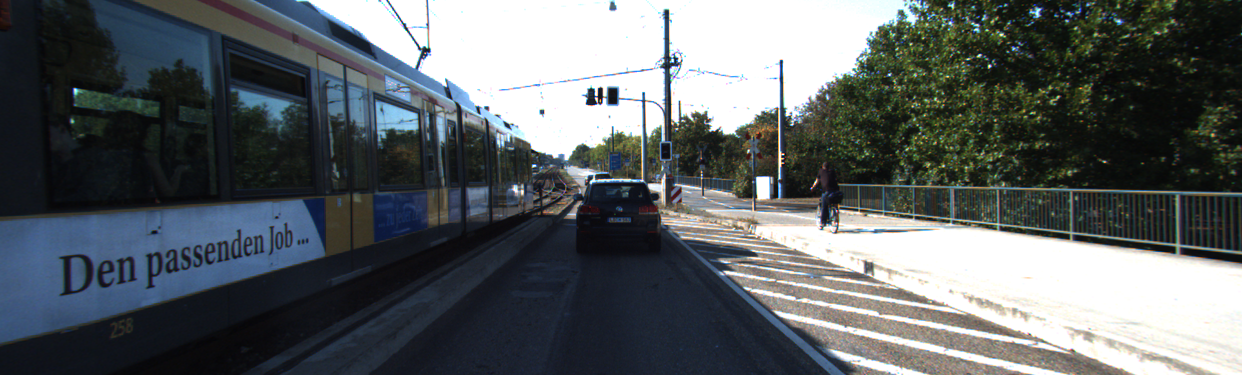}\hfill
    {(c) Input left image at 103rd frame}\\
    \includegraphics[width=\columnwidth]{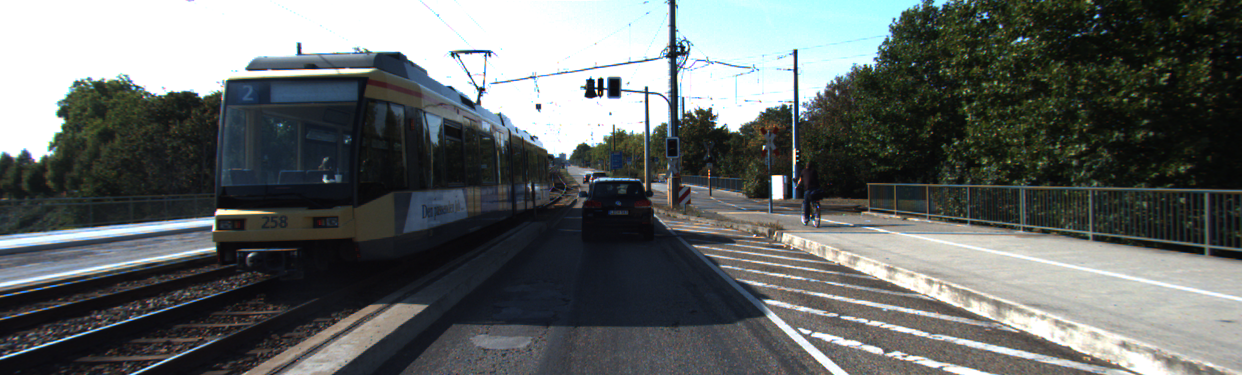}\hfill
    {(d) Input left image at 111st frame}\\
    \end{minipage}\hfill
    }\hfill
       \caption{(a) D1-all error (\%) across frames in the sequence (\textit{2011\_09\_26\_drive\_101\_sync}) from the KITTI dataset with respect to the adaptation methods. (b) - (d) are input left images at section showing high error and it seems due to sudden changes in frames.} \label{fig:1}
    \end{figure*}

    \begin{figure}[!t]
    \centering
    \renewcommand{\thesubfigure}{}
    \subfigure[(a) Base network: MADNet]{\includegraphics[width=0.5\linewidth]{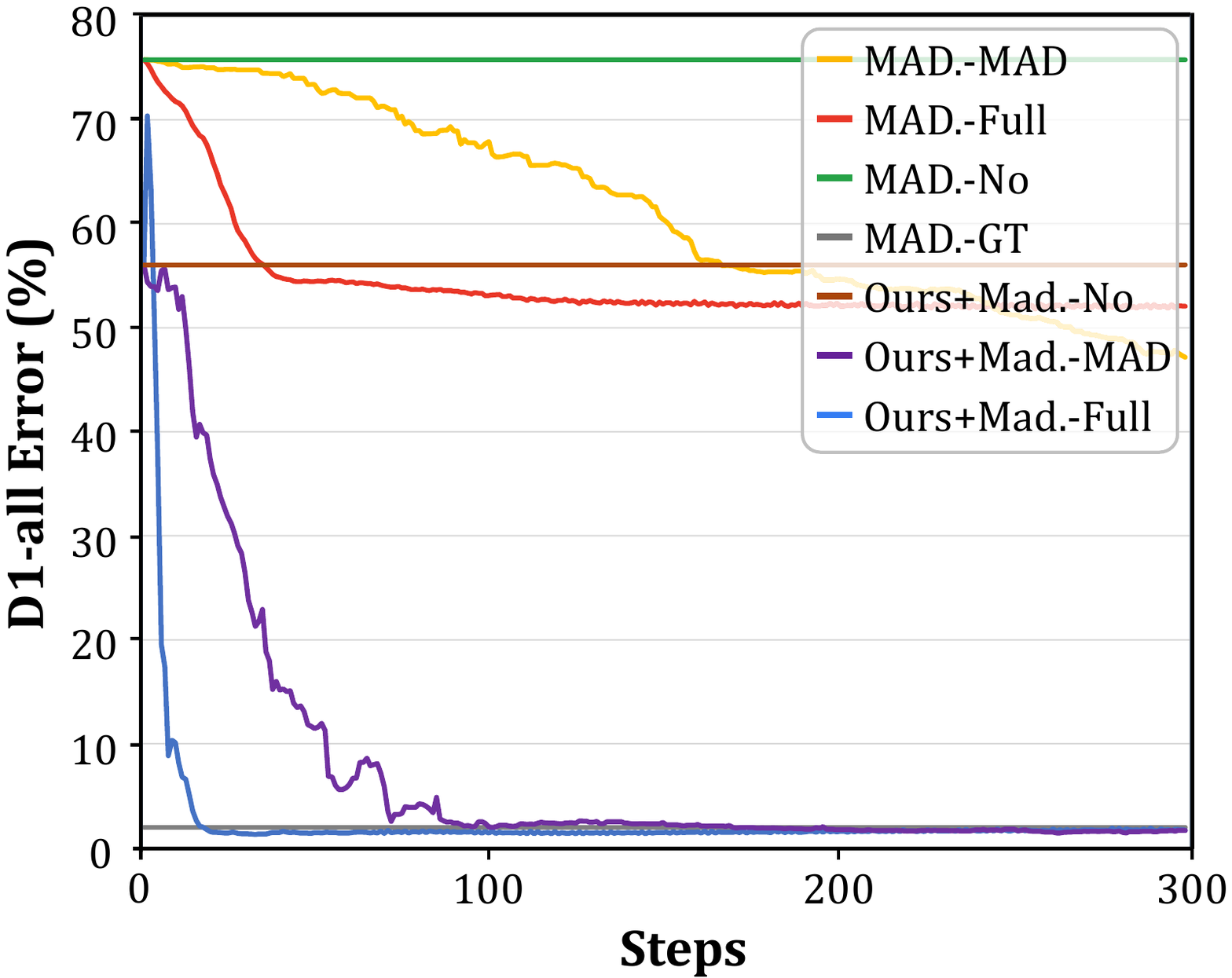}}\hfill
    \subfigure[(b) Base network: DispNetC]{\includegraphics[width=0.5\linewidth]{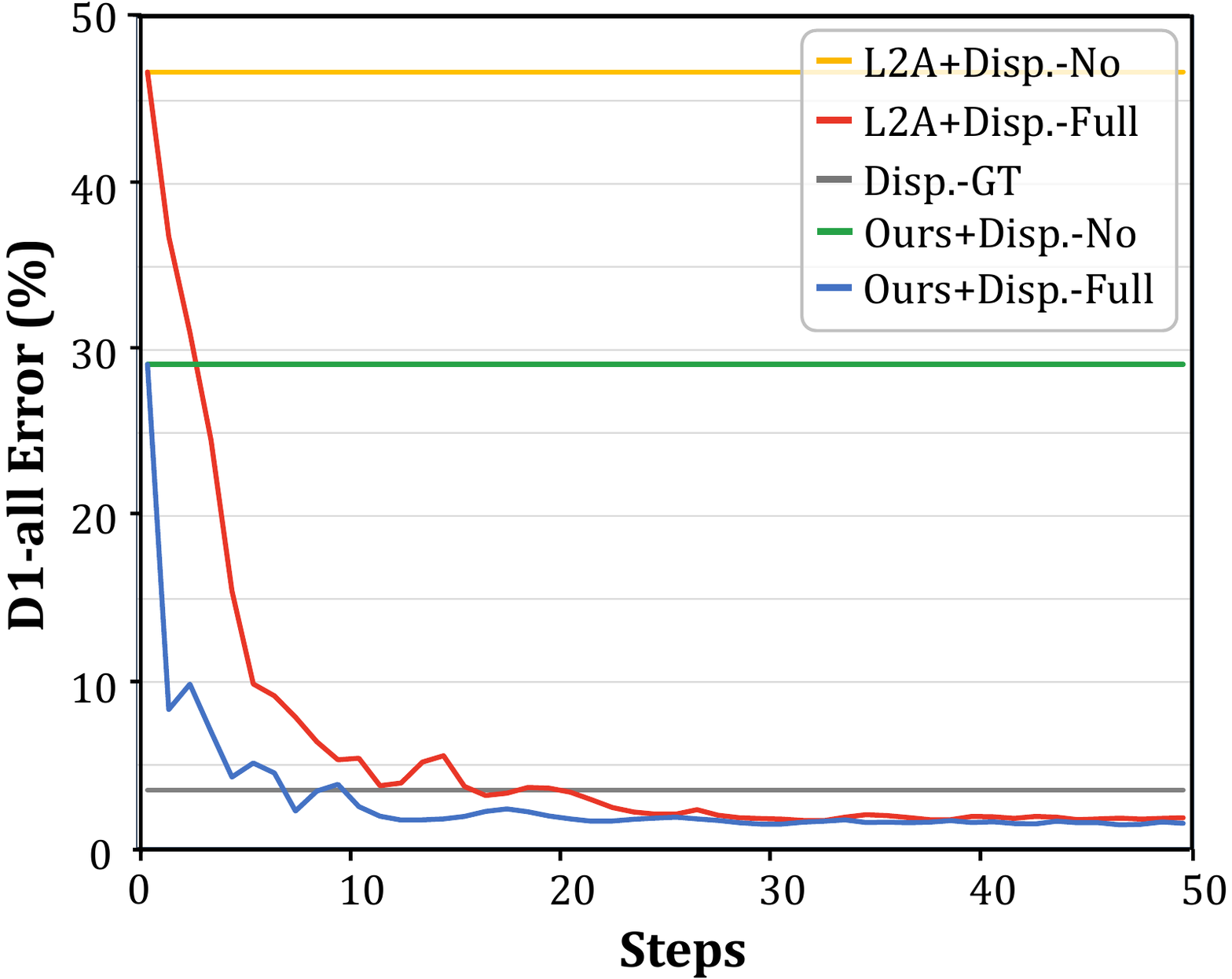}}\hfill\\ 
        \caption{Convergence analysis on a single frame with repetitive adaptation.}
    \label{fig:2}
    \end{figure}

    \begin{figure}[!t]
    \centering
    \renewcommand{\thesubfigure}{}
    \subfigure[Base network: MADNet]{\includegraphics[width=0.5\linewidth]{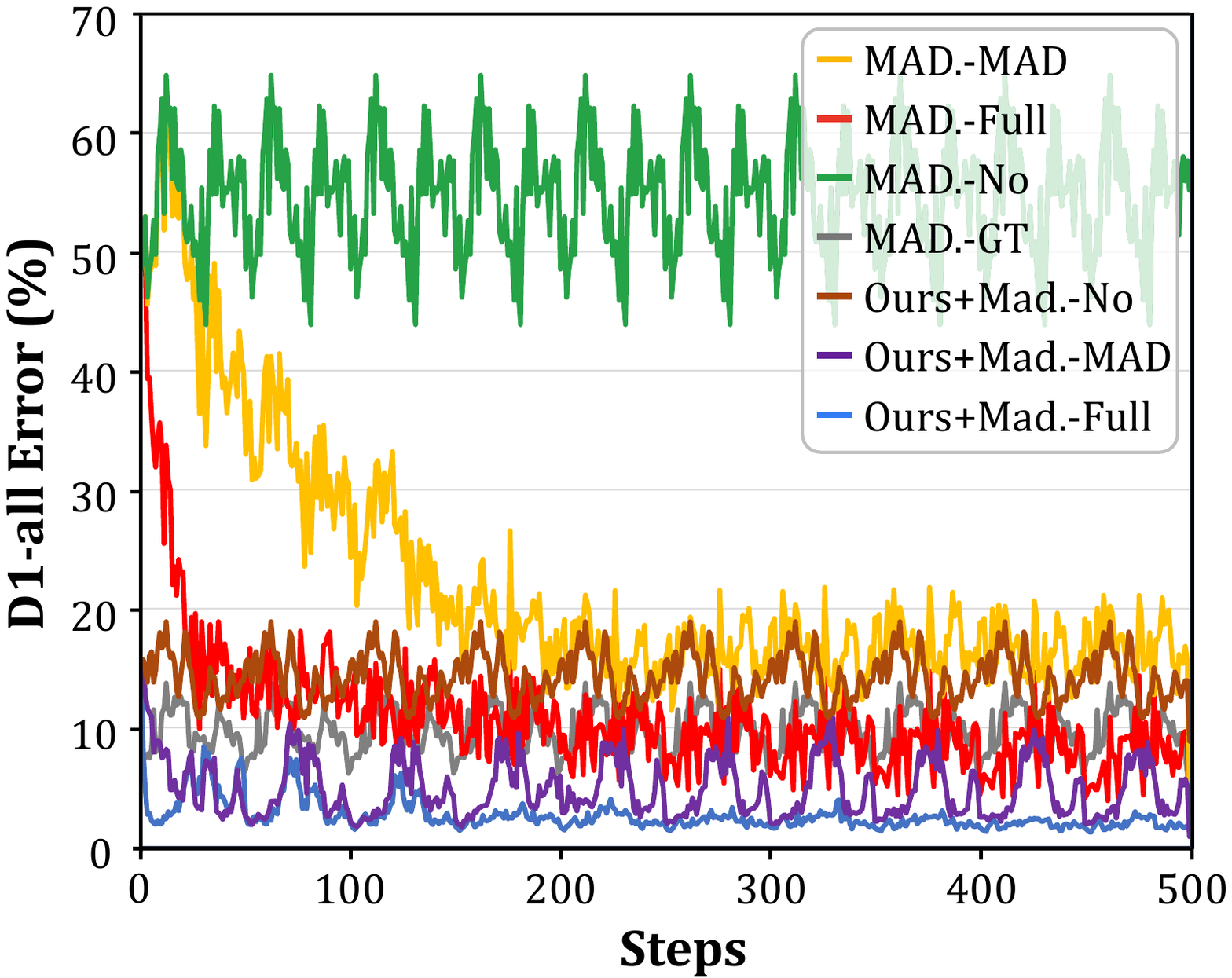}\hfill}
        \caption{Convergence analysis on a single sequence `\textit{2011\_09\_28\_drive\_0039\_sync}' with repetitive adaptation.}
    \label{fig:3}
    \end{figure}

\subsubsection{Repetitive adaptation for single frame.}
    Furthermore, our \model~takes advantage of the fast convergence.
    To validate the convergence speed, we repeatedly perform the adaptation for the first frame in a sequence\footnote{`\textit{2011\_09\_28\_drive\_0034\_sync}' sequence is used.} from the KITTI and record D1-all errors for each adaptation step according to the base network.
    In \figref{fig:2}(a), \textbf{MAD.-No} (green) and \textbf{Ours+Mad.-No} (brown) shows the performance of the initial base parameters trained using \cite{tonioni2019real} and our \model.
    From each initial parameters, the drop rate of D1-all error represents the convergence speed of the adaptation.
    With the large margin, \textbf{Ours+Mad.-Full} (blue) achieves fast convergence and outperforms \textbf{MAD.-Full} (red)~\cite{tonioni2019real}.
    The result of \textbf{Ours+Mad.-MAD} (purple) also shows fast convergence and low error compared to \textbf{MAD.-MAD} (yellow).
    For DispNetC~\cite{dispnet}, we report the results on a single frame with repetitive adaptation in \figref{fig:2}(b).
    The comparison between \textbf{L2A+Disp.-Full} (red) and \textbf{Ours+Disp.-Full} (blue) shows our \model~converges faster than L2A~\cite{tonioni2019learning}, exceeding the model fine-tuned on groundtruth (gray) with only about 10 adaptation steps.

\subsubsection{Repetitive adaptation for single sequence.}
    To conduct sequence-level convergence analysis, we perform repetitive adaptation on first 50 frames of the sequence\footnote{`\textit{2011\_09\_28\_drive\_0039\_sync}' sequence is used.} from the KITTI dataset.
    As illustrated in \figref{fig:3}, \textbf{Ours+Mad.-MAD} (purple) and \textbf{Ours+Mad.-Full} (light blue) converge with a small number of steps about 150 steps, while \textbf{MAD.-MAD} (yellow) and \textbf{MAD.-Full} (red) need about 250 steps to converge.

    \begin{table}[t]
    \centering
    \caption{Short-term and long-term adaptation performance evaluated on the Synthia~\cite{ros2016synthia} and Virtual KITTI 2~\cite{vkitti2} datasets with models trained on the F3D~\cite{dispnet} and Synthia~\cite{ros2016synthia} datasets.
    }\label{tab:syn_f3d_syn}
    \resizebox{1\linewidth}{!}{
        \begin{tabular}{lccccccccccccc}
        \hlinewd{0.9pt}
        \multirow{3}{*}{Method}    & \multirow{3}{*}{Adapt.} 
        &\multicolumn{4}{c}{F3D $\rightarrow$ Synthia} &\multicolumn{4}{c}{F3D $\rightarrow$ VKITTI2} &\multicolumn{4}{c}{Synthia $\rightarrow$ VKITTI2} \tabularnewline \cline{3-6} \cline{7-10} \cline{11-14} & 
        & \multicolumn{2}{c}{Short-term} &  \multicolumn{2}{c}{Long-term}
        & \multicolumn{2}{c}{Short-term} &  \multicolumn{2}{c}{Long-term}
        & \multicolumn{2}{c}{Short-term} &  \multicolumn{2}{c}{Long-term}
        \\ \cline{3-4}  \cline{5-6} \cline{7-8} \cline{9-10} \cline{11-12} \cline{13-14} &   
        & D1-all & EPE & D1-all & EPE & D1-all & EPE & D1-all & EPE & D1-all & EPE & D1-all & EPE \tabularnewline
        \hline \hline
        MADNet           &No    &48.49  &20.86  &48.00  &20.39  &47.91  &17.62  &47.13  &17.09  &46.95  &13.36  &42.29  &9.30    \tabularnewline
        Ours-MAD.        &No    &29.97  &5.00   &29.96  &5.01  &27.33   &3.19   &26.50  &3.15   &28.96  &4.93   &27.02  &3.93   \tabularnewline
        MADNet           &Full  &21.02  &5.10   &17.23  &3.27  &25.69   &3.83    &18.38  &2.57  &22.47  &3.31   &\textbf{17.70}  &2.59   \tabularnewline
        Ours-MAD.        &Full  &\textbf{18.12}  &\textbf{2.97}   &\textbf{14.03}  &\textbf{2.67}  &21.61   &\textbf{2.72}   &18.06  &2.56   &18.27  &2.42   &17.73  &2.50 \tabularnewline 
        MADNet           &MAD   &19.98  &4.12   &18.20  &3.58  &23.90   &3.12   &18.55  &2.62   &29.29  &5.53   &18.66  &2.55   \tabularnewline
        Ours-MAD.        &MAD   &22.55  &4.03   &18.27  &3.48  &\textbf{21.49}   &2.78   &\textbf{17.49}  &\textbf{2.49}   &\textbf{16.90}   &\textbf{2.24}   &18.11  &\textbf{2.47}  \tabularnewline \hline

        \hlinewd{0.9pt}
        \end{tabular}
        }
    \end{table}
    
    \begin{table*}[t]
    \centering
    \caption{Mid-term adaptation on the Virtual KITTI 2~\cite{vkitti2} with models trained on the Synthia~\cite{ros2016synthia} dataset.
    }\label{tab:vkitti-mid-range}
    \resizebox{\linewidth}{!}{
        \begin{tabular}{@{\extracolsep{4pt}}lccccccccccccccc@{}}
        \hline
        \multirow{2}{*}{Method} & \multirow{2}{*}{Adapt.}  
        & \multicolumn{2}{c}{clone}  & \multicolumn{2}{c}{fog}             
        & \multicolumn{2}{c}{morning} & \multicolumn{2}{c}{overcast}
        & \multicolumn{2}{c}{rain} & \multicolumn{2}{c}{sunset}
        & \multicolumn{2}{c}{Avg.}  
        \\ \cline{3-4}  \cline{5-6} \cline{7-8} \cline{9-10} \cline{11-12} \cline{13-14} \cline{15-16}
        & 
        & \multicolumn{1}{l}{D1-all } & \multicolumn{1}{l}{EPE} 
        & \multicolumn{1}{l}{D1-all } & \multicolumn{1}{l}{EPE} 
        & \multicolumn{1}{l}{D1-all } & \multicolumn{1}{l}{EPE} 
        & \multicolumn{1}{l}{D1-all } & \multicolumn{1}{l}{EPE} 
        & \multicolumn{1}{l}{D1-all } & \multicolumn{1}{l}{EPE} 
        & \multicolumn{1}{l}{D1-all } & \multicolumn{1}{l}{EPE} 
        & \multicolumn{1}{l}{D1-all } & \multicolumn{1}{l}{EPE} 
        \\ \hline \hline
        MADNet      & No                     
        &33.25  &7.90   &69.62  &14.57  &33.86  &7.60   &34.25  &7.92   &48.38  &9.72   &34.39  &8.07   
        &42.29  &9.30   \\
        Ours-MAD.   & No
        &23.16  &3.42   &32.71  &4.65   &25.50  &3.92   &24.07  &3.53   &31.78  &4.43   &24.88  &3.65
        &27.02  &3.93   \\    \hline
        MADNet      & Full  
        &12.98  &1.83   &25.08  &3.11   &13.63  &2.11   &21.91  &2.67   &30.46  &4.13   &14.84  &2.12   
        &19.82  &2.66   \\
        Ours-MAD.   & Full  
        &14.69  &2.05   &24.16  &3.00   &13.25  &2.10   &20.23  &2.63   &27.92  &3.92   &14.54  &2.20
        &19.13  &2.65   \\    \hline
        MADNet      & MAD                     
        &19.32  &2.58   &32.03  &4.03   &16.56  &2.33   &23.74  &2.92   &30.79  &4.26   &18.96  &2.44
        &23.57  &3.09   \\
        Ours-MAD.   & MAD
        &\textbf{12.74}  &\textbf{1.71}   &\textbf{23.48}  &\textbf{2.95}   &\textbf{12.69}  &\textbf{1.88}   &\textbf{14.85}  &\textbf{1.85}   &\textbf{25.63}  &\textbf{3.17}   &\textbf{13.10}  &\textbf{1.82}
        &\textbf{17.08}  &\textbf{2.23}   \\    \hlinewd{0.8pt}
    \end{tabular}
    }
    \end{table*}

    \subsection{Synthetic to Synthetic Adaptation.}\label{sec:32}
    To verify the general adaptability of the proposed method in various scenarios that are not limited to synthetic-to-real, we further evaluate the performance under the short-, mid-, and long-term adaptation settings on several synthetic driving scene benchmarks.
    We present results according to the adaptation methods in \tabref{tab:syn_f3d_syn} and \tabref{tab:vkitti-mid-range}.
    
    In \tabref{tab:syn_f3d_syn}, we train the models on the F3D~\cite{dispnet} or Synthia~\cite{ros2016synthia} datasets and evaluate the short- and long-term adaptation performance on the Synthia~\cite{ros2016synthia} or Virtual KITTI 2~\cite{vkitti2} datasets.
    Note that each sequence for the short-term adaptation is defined as a distinct sequence provided from each dataset (\eg \textit{SYNTHIA-SEQS-01-DAWN} in the Synthia) and all frames are concatenated into a single sequence for the long-term adaptation.
    The comparison between the MADNet~\cite{tonioni2019real} and Ours-MAD. with \textit{No adaptation} shows the initial parameters of our \model~are extremely powerful to the new domain in terms of D1-all error and EPE on all datasets.
    The state-of-the-art performances using the adaptation are attained by our method except for one metric (\ie D1-all error of the long-term adaptation on Synthia$\rightarrow$VKITTI2).
    The results evaluated on the Virtual KITTI 2 dataset indicates that the performances are different according to the training data.
    Since both the Synthia and Virtual KITTI 2 are driving scene datasets, the higher performance can be obtained when the models are trained on the Synthia than F3D.

    To verify the robustness of our method under the mid-term adaptation setting, we evaluate the performance on various weather or lighting conditions using the Virtual KITTI 2~\cite{vkitti2} dataset.
    Therefore, we provide the adaptation performance according to 6 different conditions and averaged performance.
    As shown in \tabref{tab:vkitti-mid-range}, our method consistently surpasses the baseline in all conditions, outperforming the MADNet with a large margin.
    Although the performance on the poor weather conditions (\eg fog or rain) that have highly different pixel distributions from the training data shows relatively high error, the proposed method still outperforms the MADNet by $8.55\%$ and $5.16\%$ in terms of D1-all error, and $1.08$ and $1.09$ in terms of EPE for fog and rain sequences, respectively.

\subsection{More qualitative results.}\label{sec:33}

\begin{figure}[!th]
       \centering
       \subfigure{
    \begin{minipage}{0.52\linewidth}\centering
    \includegraphics[width=\columnwidth]{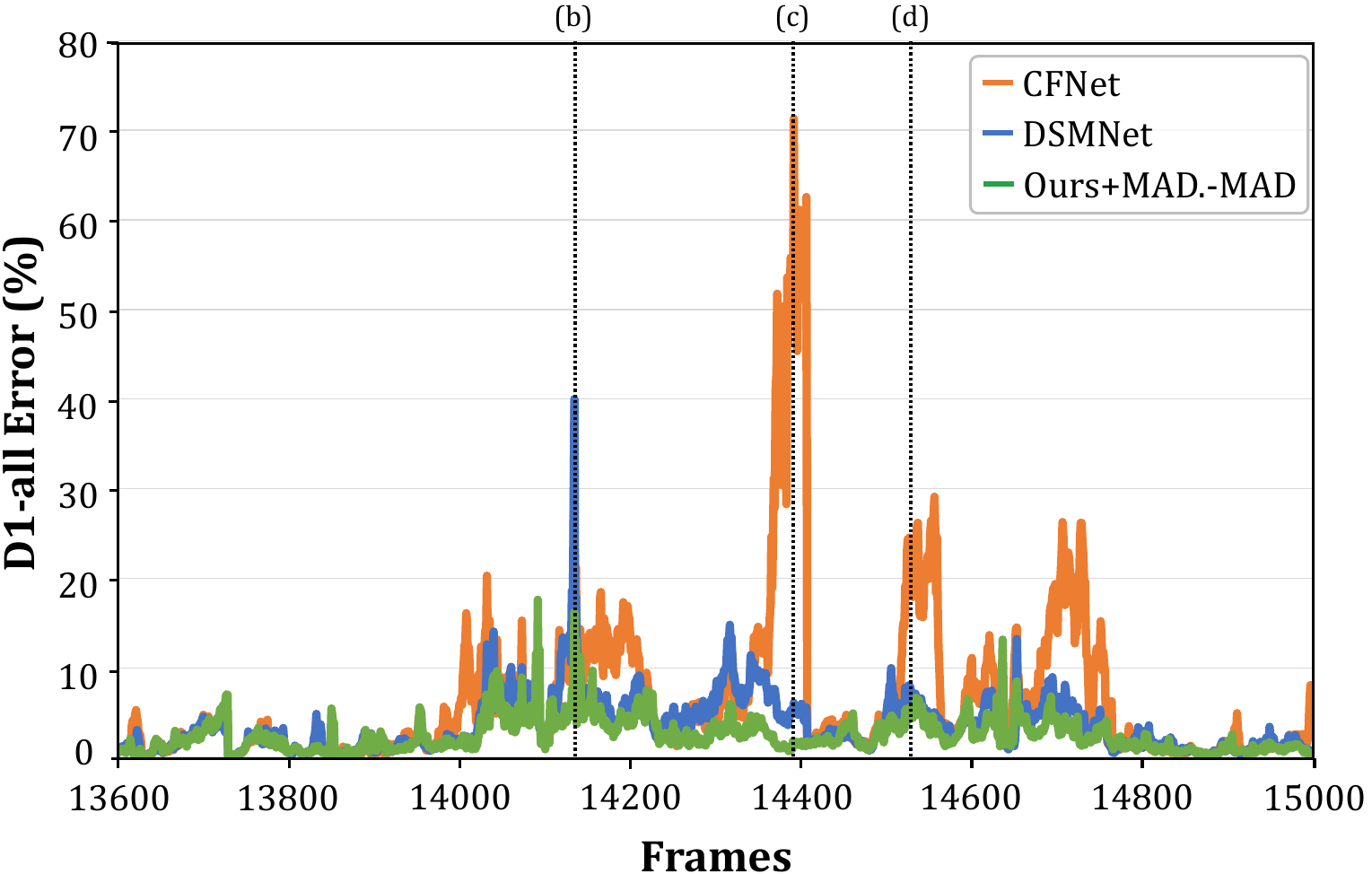}\hfill
    {(a) D1-all error (\%) across frames}\\
    \end{minipage}\hfill
    }\hfill
    \subfigure{
    \begin{minipage}{0.45\linewidth}\centering
    \includegraphics[width=\columnwidth]{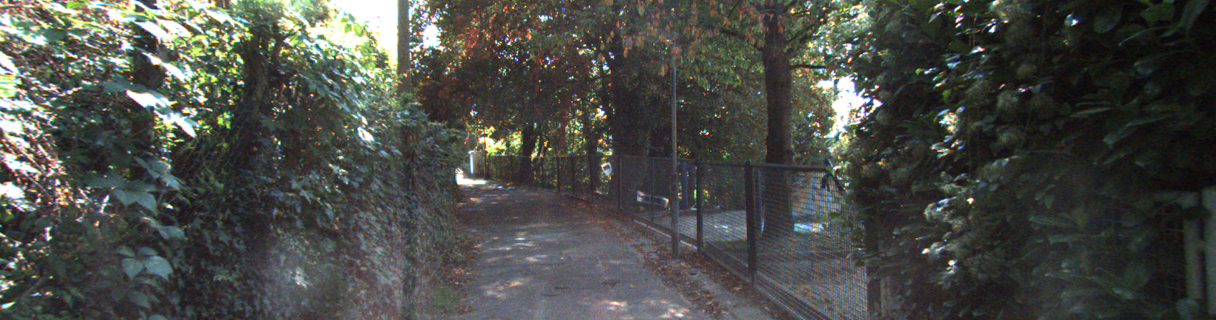}\hfill
    {(b) Input left image at 14150th frame}\\
    \includegraphics[width=\columnwidth]{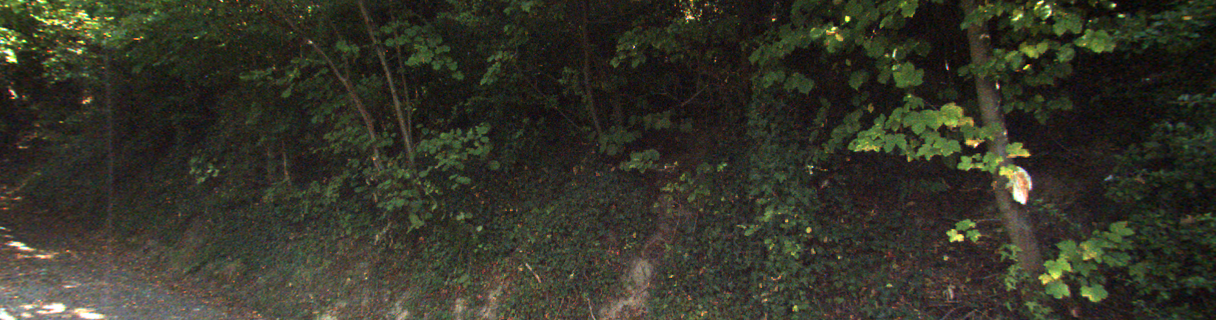}\hfill
    {(c) Input left image at 14414th frame}\\
    \includegraphics[width=\columnwidth]{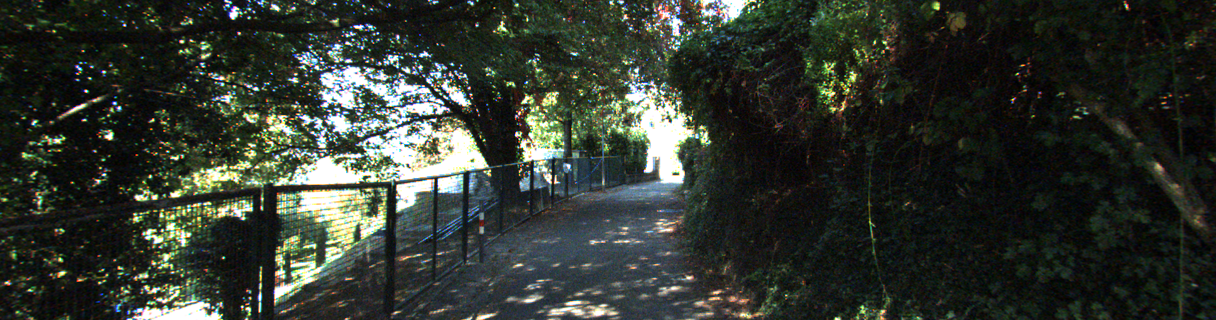}\hfill
    {(d) Input left image at 14551st frame}\\
    \end{minipage}\hfill
    }\hfill
       \caption{(a) D1-all error (\%) across frames in the consecutive sequences including `\textit{2011\_09\_26\_drive\_0086\_sync}' (frames in (b) and (c)) and `\textit{2011\_09\_26\_drive\_0087\_sync}' (frame in (d)) from the KITTI~\cite{geiger2013vision} dataset with respect to the adaptation methods. (b) - (d) are input left images at section showing high error and it seems due to sudden changes in frames.} \label{fig:dg}
    \end{figure}

    To argue the necessity of the online adaptation, we present the qualitative results of the proposed method compared to domain generalization (DG) approaches~\cite{dsmnet,cfnet}.
    We evaluate state-of-the-art DG models~\cite{dsmnet,cfnet} for all frames in the KITTI~\cite{geiger2013vision} dataset.
    In this experiment, we use the MADNet as the base network and perform the long-term adaptation.
    \figref{fig:dg}-(a) shows the performance in terms of D1-all error over all frames from the KITTI dataset.
    For the visibility of the results, we depict the performance corresponding to the $13600\sim15000$th frame.
    As shown in the figure, \textbf{Ours-MAD.} (green) achieves lower error rate than \textbf{DSMNet}~\cite{dsmnet} (blue) and \textbf{CFNet}~\cite{cfnet} (orange).
    In addition, we depict input left frames showing high error rates on the DG methods, as shown in \figref{fig:dg}(b)-(d).
    Although the scene is not abruptly changed or the intensity of illumination is not drastically changed, the DG methods often show remarkably low performance compared with our method, which has stable performance in overall frames.
    Thus, the fast inference speed and robust performance in novel environments make our method more practical to real-world applications\footnote{Analysis for the inference speed is provided in the main paper.}.

\begin{figure*}[t]
    \begin{center}
       \includegraphics[width=1\linewidth]{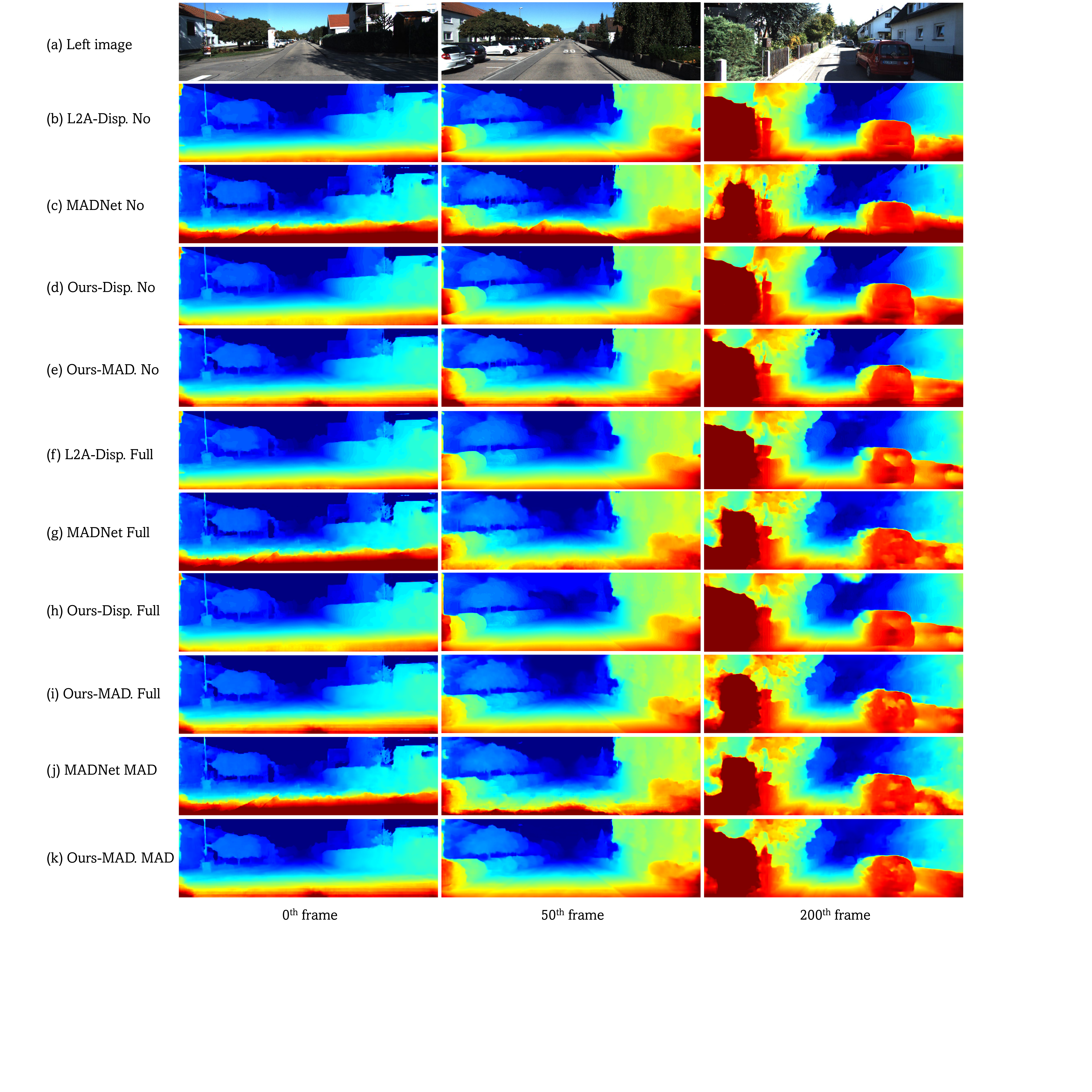}
    \end{center}
       \caption{Disparity maps predicted using MADNet and DispNet as the base network on the KITTI sequence~\cite{geiger2013vision}. (a) Input left images; predicted disparity with (b)-(e) no adaptation; (f)-(i) full adaptation; and (j), (k) MAD adaptation.}
    \label{fig:4}
\end{figure*}

    In \figref{fig:4}, we present more qualitative results evaluated on the sequence from the KITTI dataset\footnote{`\textit{2011\_09\_28\_drive\_0018\_sync}' sequence is used.} according to the base network and the adaptation method.
    The results contain the quality of the initialized parameters(first column), fast adaptation (second column), and convergence to low errors (last column).
    For the comparison between methods with no adaptation (b)-(e), our \model~shows visually better results at the inner or boundary of objects regardless of the base network.
    The comparison between the results in the second column, $\mathbf{L2A-Disp.Full}$ (f), $\mathbf{Ours-Disp.Full}$ (h), $\mathbf{Ours-MAD.Full}$ (i), and $\mathbf{Ours-MAD.MAD}$ (k) show the superior performance with only 50 adaptation steps.
    The results in the third column, our \model~converges to the low error, showing clear prediction on the inner and boundary of objects.

\subsection{Additional ablation studies}\label{sec:addabl}

\begin{table}[t]
\small
    \centering
    \caption{Ablation studies for various components. All results are obtained using \textit{Full adaptation} on short-term sequences.
    }\label{tab:abl}
    {
        \begin{tabular}{lccc}
        \hline
        Method   & D1-all & EPE \tabularnewline
        \hline \hline
        No PointFixNet (w/ selection)    &7.77  &2.02 \tabularnewline
        No residual learning            &2.10   &0.82  \tabularnewline
        Point selection threshold 1     &\textbf{1.97}   &0.80  \tabularnewline
        Point selection threshold 5     &2.01   &0.80  \tabularnewline 
        No online ML                    &5.53   &1.13  \tabularnewline
        Ours-MAD.                       &2.00   &\textbf{0.74} \tabularnewline \hline
        
        \end{tabular}
    }
\end{table}

\subsubsection{Ablation study on \model Net.}
To further verify the effectiveness of \model Net, we ablate \model Net while keeping the point selection process. As shown in \tabref{tab:abl}(row 1), the model trained using the sparse loss without \model Net records the poor performance.

\subsubsection{Ablation study on residual learning.}
We adopted residual learning to make back-propagation flow through skip connection. The performance without residual learning shown in \tabref{tab:abl}(row 2) degrades slightly compared to model trained with residual learning(row 6).

\subsubsection{Ablation study on point selection threshold.}
We choose 3 pixels error as threshold for point selection motivated by the \textit{bad3} metric. As shown in \tabref{tab:abl}(row 3,4 and 6), the best performance in terms of EPE is attained with 3 pixels.

\subsubsection{Ablation study on online meta-learning.}
We adopted online meta-learning framework to jointly train \model Net and the base network. The performance of offline meta-learning is shown in \tabref{tab:abl}(row 5) indicating the advantage of online meta-learning in our framework.

\clearpage

\clearpage

\end{document}